\documentclass[letterpaper]{article} 
\usepackage{aaai2026}  
\usepackage{times}  
\usepackage{helvet}  
\usepackage{courier}  
\usepackage[hyphens]{url}  
\usepackage{graphicx} 
\usepackage{natbib}  
\usepackage{caption} 
\usepackage{amsmath}
\usepackage{algorithmic}
\usepackage{algorithm}
\usepackage{array}
\usepackage{textcomp}
\usepackage{url}
\usepackage{verbatim}
\usepackage{graphicx}
\usepackage{cite}
\usepackage{amssymb}
\usepackage{booktabs}
\usepackage{enumitem}
\usepackage{times}
\usepackage{epsfig}
\usepackage{amsfonts}
\usepackage{bm}
\usepackage{multirow}
\usepackage{makecell}
\usepackage{bbding}
\usepackage[table]{xcolor}
\usepackage{tikz}
\usepackage{subcaption}
\usepackage{pifont}

\newcommand{\ignore}[1]{}
\definecolor{cGreen}{RGB}{100,180,100}
\definecolor{cRed}{RGB}{220,50,0}

\title{Tracking and Segmenting Anything in Any Modality}
\author{
    Tianlu Zhang\textsuperscript{\rm 1},
    Qiang Zhang\textsuperscript{\rm 2},
    Guiguang Ding\textsuperscript{\rm 3},
    Jungong Han\textsuperscript{\rm 1}
    \thanks{Corresponding author: jghan@tsinghua.edu.cn.}
}
\affiliations{
    \textsuperscript{\rm 1}Department of
    Automation, Tsinghua University\\
    \textsuperscript{\rm 2}School of Mechano-Electronic Engineering, Xidian University\\
    \textsuperscript{\rm 3}School of Software, Tsinghua University.\\
}

\usepackage{bibentry}

\begin{document}

\maketitle

\begin{abstract}
Tracking and segmentation play essential roles in video understanding, providing basic positional information and temporal association of objects within video sequences. 
Despite their shared objective, existing approaches often tackle these tasks using specialized architectures or modality-specific parameters, limiting their generalization and scalability.
Recent efforts have attempted to unify multiple tracking and segmentation subtasks from the perspectives of any modality input or multi-task inference.
However, these approaches tend to overlook two critical challenges: the distributional gap across different modalities and the feature representation gap across tasks. These issues hinder effective cross-task and cross-modal knowledge sharing, ultimately constraining the development of a true generalist model.
To address these limitations, we propose a universal tracking and segmentation framework named SATA, which unifies a broad spectrum of tracking and segmentation subtasks with any modality input. 
Specifically, a Decoupled Mixture-of-Expert (DeMoE) mechanism is presented to decouple the unified representation learning task into the modeling process of cross-modal shared knowledge and specific information, thus enabling the model to maintain flexibility while enhancing generalization.
Additionally, we introduce a Task-aware Multi-object Tracking (TaMOT) pipeline to unify all the task outputs as a unified set of instances with calibrated ID information, thereby alleviating the degradation of task-specific knowledge during multi-task training.
SATA demonstrates superior performance on 18 challenging tracking and segmentation benchmarks, offering a novel perspective for more generalizable video understanding.
\end{abstract}

\section{Introduction}

Video, as a crucial medium for information dissemination, has emerged as a dominant constituent of internet traffic. This prevalence has notably spurred the advancement of video understanding technologies. 
At present, video understanding has witnessed substantial expansion and now encompasses a wide range of tasks, including action recognition, video segmentation, object tracking, and etc. Among these, object tracking and segmentation are dedicated to establishing instance-level or pixel-level correspondences across frames, thereby laying the groundwork for tackling video tasks. 

\begin{figure}[t]
\centering
\includegraphics[width=0.95\columnwidth]{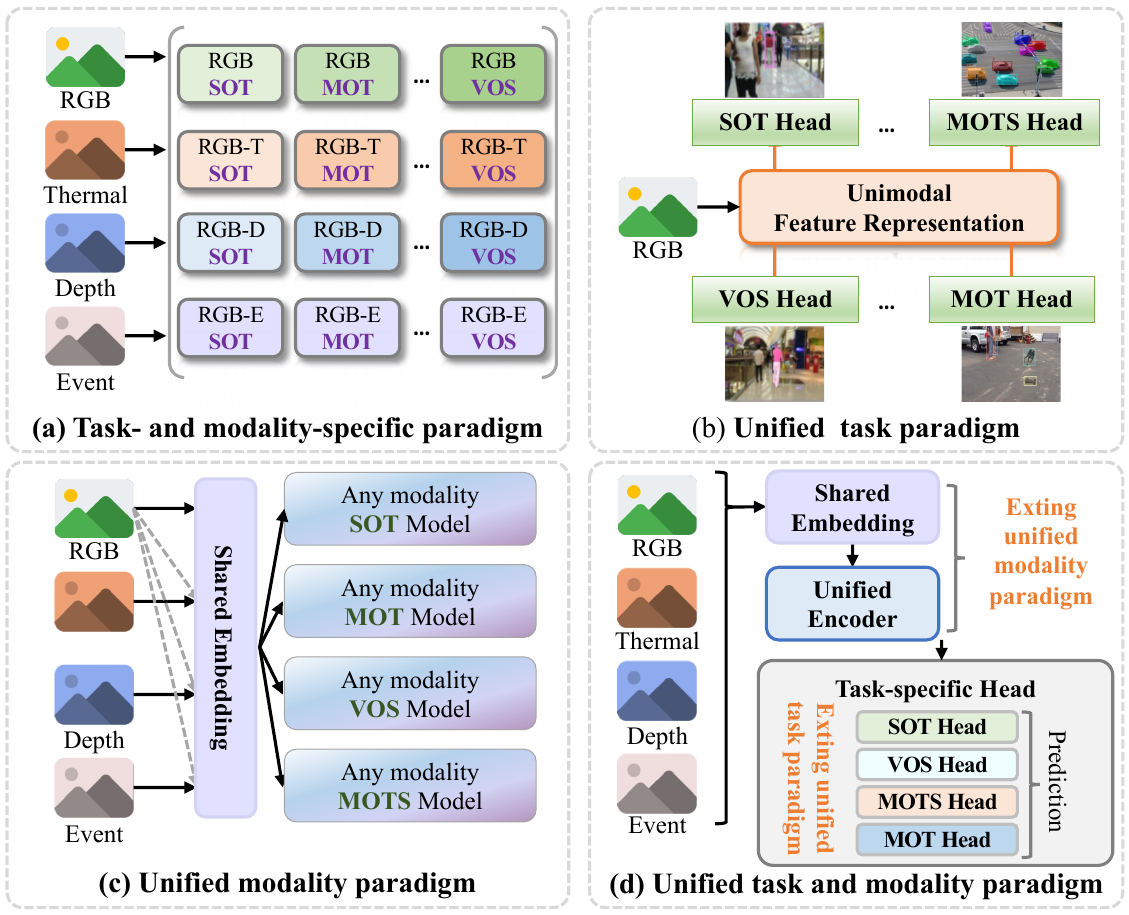}
\caption{Illustration of existing tracking and segmentation paradigm. (a) Task- and modality-specific paradigm. (b) Unified task paradigm. (c) Unified modality paradigm. (d) Unified task and modality paradigm obtained by combining existing unified task methods and unified modality models.
}
\label{fig1}
\vspace{-0.4cm}
\end{figure}

Over the years, the object tracking and segmentation have developed into multiple subtask branches, including Single Object Tracking (SOT)~\cite{got10k}, Multiple Object Tracking (MOT)~\cite{masa}, Video Object Segmentation (VOS)~\cite{sam2} and Multi-Object Tracking and Segmentation (MOTS)~\cite{uninext}.
In addition to the widely used RGB cameras, various sensors have been introduced to enhance the performance of tracking and segmentation in complex scenarios, e.g., depth, thermal, and event data.
For a long period, research in these typical tracking and segmentation subtasks has adopted a task- and modality-specific paradigm, i.e., designing specialized architectures and loss functions to cater to the unique requirements of different subtasks (e.g., SOT, MOT and VOS) and fixed input modalities (e.g., RGB, RGB-T and RGB-D), as shown in Fig.~\ref{fig1} (a). 
The divergent setups require customized methods with carefully designed architectures and hyper-parameters, leading to complex training and redundant parameters. 

To mitigate this issue, some recent works have explored the possibility of establishing general models from two perspectives: unified task paradigm and unified modality paradigm.
Specifically, as shown in Fig.~\ref{fig1} (b), the unified task paradigm~\cite{uninext,sutrack,omnivid} breaks through the isolation of task domain knowledge and general knowledge, establishing the implicit collaboration between different tasks.
Meanwhile, as shown in Fig.~\ref{fig1} (c), the unified modality methods~\cite{untrack,sutrack} consolidate the input of any modality using a unified model with shared parameters.
The above two methods respectively establish unified modality representation learning and unified task feature space modeling, thereby reducing the complexity of model design and the need for extensive hyper-parameter tuning across various input modalities and sub-task combinations.

Recently, due to the potential to realize Artificial General Intelligence (AGI), these unified models have drawn great attention. 
A natural idea is: combining the above two paths to achieve a more generalizable video tracking and segmentation model capable of handling any input modality and supporting multi-task inference, as shown in Fig.~\ref{fig1} (d).
While this strategy is conceptually straightforward and partially effective, it overlooks the distributional discrepancies across modalities and the representational gaps between tasks—factors that limit the generalist model’s ability to fully leverage cross-modal and cross-task knowledge.

\begin{figure}[t]
\centering
\includegraphics[width=0.85\columnwidth]{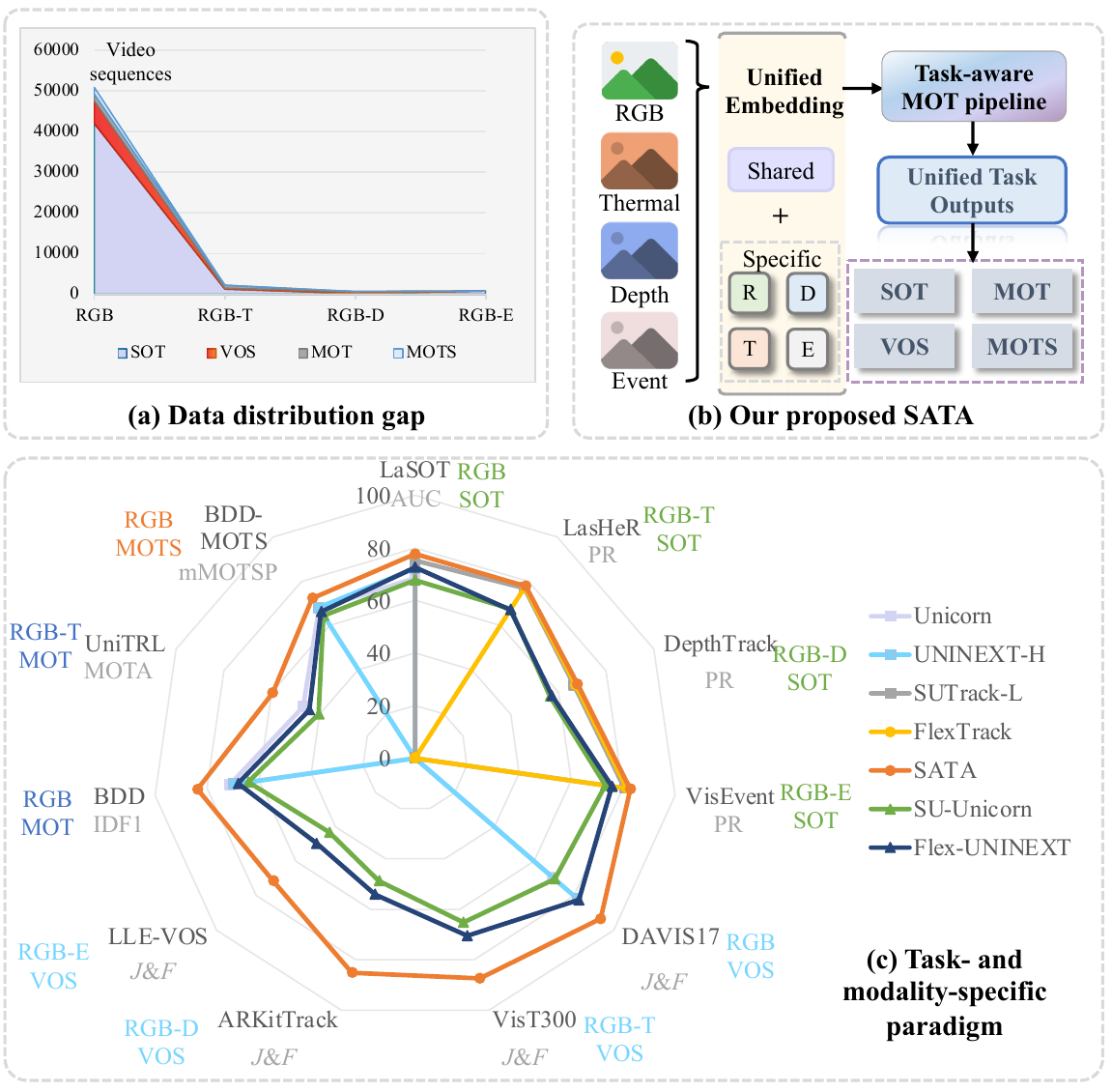}
\caption{Analysis of data distribution gap and comparison of the proposed SATA with existing strategies.
(a) Statistical overview of the data distribution gap during model training.
(b) Overview of the proposed SATA framework.
(c) Our SATA v.s. the existing methods on 11 challenging benchmarks. Here, SU-Unicorn denotes the combination of SUTrack~\cite{sutrack} and Unicorn~\cite{unicorn}, Flex-UNINEXT denotes the combination of FlexTrack~\cite{FlexTrack} and UNINEXT~\cite{uninext}.
}
\label{fig2}
\vspace{-0.3cm}
\end{figure}

Specifically, the distribution gap in multi-modal data arises not only from discrepancies in cross-modal information but also from differences in their underlying representations.
Most existing unified modality approaches~\cite{sutrack,FlexTrack} have attempted to learn a cohesive embedding across diverse input modalities, but the effectiveness of modality-specific clues have not been fully exploited.
Similarly, while generic representations can be learned with the unified task approaches, they are still constrained by elaborated-designed task-specific heads and isolated multi-stage training strategies across different downstream tasks~\cite{unicorn,uninext,omnivid}. 
These fragmented architectures and learning paradigms restrict the model’s ability to acquire truly generalizable knowledge. Moreover, as shown in Fig.~\ref{fig2} (a), inconsistencies in the quality and scale of training data across different multi-modal subtasks further exacerbate task and modality biases in unified models.

With this in mind, we present SATA, a unified tracking and segmentation framework that models a unified representation of arbitrary modalities and approaches a broad spectrum of tracking and segmentation subtasks as MOT inference task conditioned on different priors, as shown in Fig.~\ref{fig2} (b).
Technically, we introduce a Decoupled Mixture-of-Expert (DeMoE) mechanism to decouple the unified representation learning into parallel modeling of modality-common and modality-specific knowledge, thus achieving a more comprehensive representation of any modality input.
In addition, unlike previous unified task models that employ several task-specific heads, we present a Task-aware MOT pipeline to unify all the task outputs (i.e., SOT, VOS, MOT, MOTS) as a unified set of instances with calibrated ID information, thereby alleviating the degradation of task-specific knowledge during multi-task training.
With the unified model architecture and totally-shared parameters, SATA can solve 4 type of tracking and segmentation subtasks of 4 combinations of input modality.
As shown in Fig.~\ref{fig2} (c), our proposed SATA significantly outperforms the simple combination of  existing unified methods on all downstream tasks.
We summarize that our work has the following contributions:

\begin{itemize}
    \item To the best of knowledge, SATA is the first unified framework capable of performing both tracking and segmentation tasks in arbitrary modality input and multi-task joint prediction.
    \item We propose a Decoupled Mixture-of-Expert (DeMoE) mechanism and a Task-aware MOT pipeline to address the distribution gap in multi-modal data and the feature representation gap across tasks, enabling more effective cross-modality and cross-task knowledge sharing.
    \item We show superior results of our proposed method on 18 challenging benchmarks from 4 subtasks tasks, all using the same model architecture and parameter set.
\end{itemize}

\section{Related works}

\begin{figure*}[t]
\centering
\includegraphics[width=0.75\linewidth]{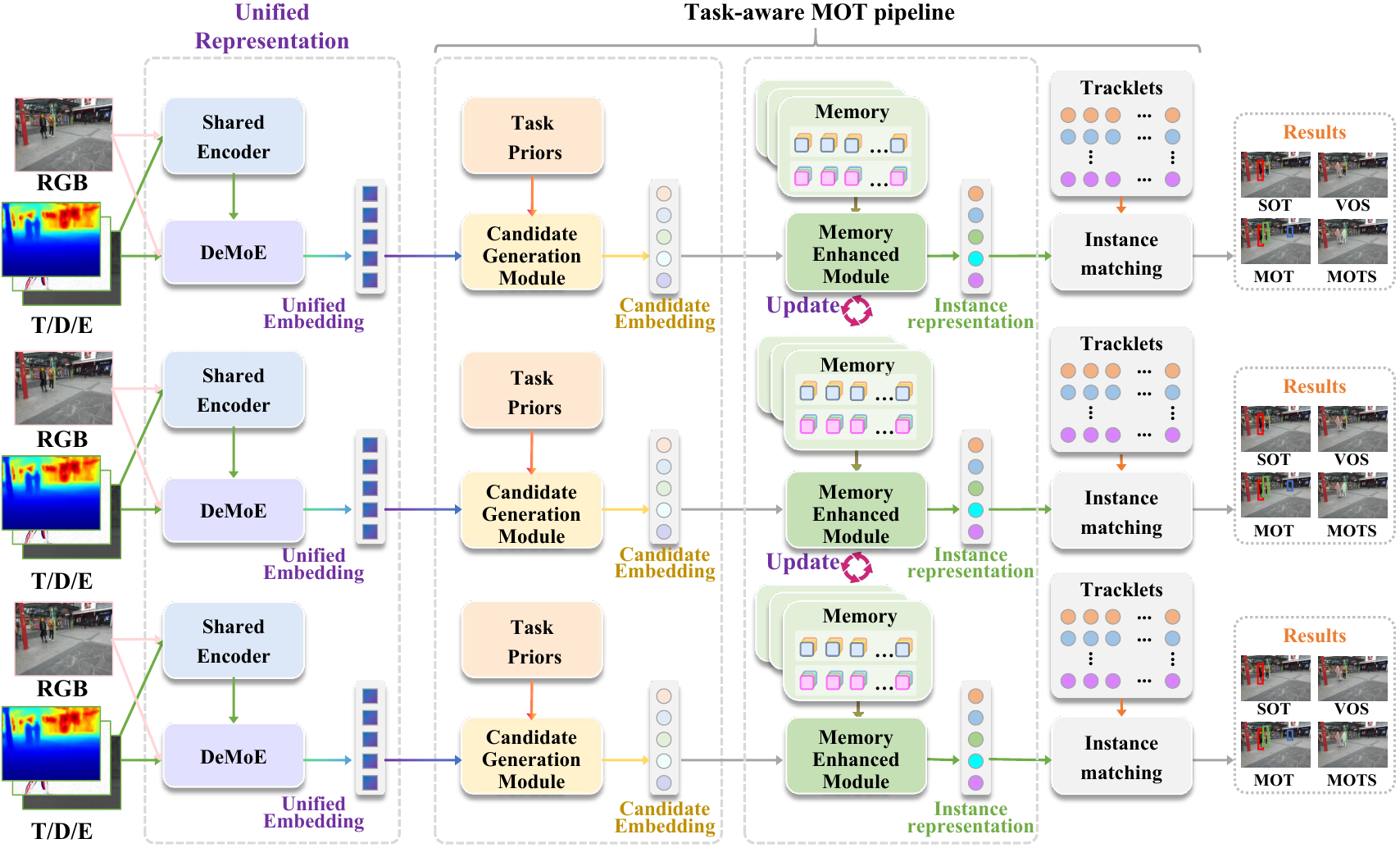}
\caption{Overview architecture of our SATA, which consists of two core components: the Decoupled Mixture-of-Expert mechanism and the Task-aware MOT pipeline.}
\label{fig3}
\vspace{-0.5cm}
\end{figure*}

\textbf{Task- and modality-specific Methods.}
Over the years, a wide variety of task-specific models have been proposed to continuously improve the tracking and segmenting performance in specific scenarios, e.g., SOT, MOT, VOS and MOTS.
SOT~\cite{got10k} and VOS~\cite{sam2} specify tracked objects on the first frame of a video using boxes or masks, then require algorithms to predict the trajectories of the tracked objects in boxes or masks, respectively. 
At present, Transformers based methods have mainstream SOT, which can be categorized into classification- and regression-based~\cite{ostrack,lorat}, corner prediction-based~\cite{mixformer,mixformerv2}, and sequence-learning-based trackers~\cite{seqtrack}. 
Meanwhile, memory-based VOS approaches have become the dominant method in VOS, which extend a single vision transformer to jointly process the current frame along with all previous frames and associated predictions~\cite{sam2, xmem}.
Different from SOT and VOS, MOT and MOTS aim to find and associate all instances~\cite{unicorn}.
The mainstream methods follow the tracking-by-detection paradigm~\cite{unicorn} and tracking-by-query~\cite{transmot} pipeline. 
With the development of sensor technology, tracking and segmentation tasks have evolved from using only a RGB camera to introduce various auxiliary modalities~\cite{sutrack,dmtrack,transcmd}.

\textbf{Unified Modality Methods.}
Despite the success of previous multi-modal methods tailored for modality-specific input, the inherent multi-parameter set paradigm limits the flexibility of these models in practical application scenarios, and generally requires the deployment of multiple models for different situations, greatly increasing the training and deployment costs.
To echo this problem, some works~\cite{vipt, onetracker, sdstrack, cstrack} adopt a architecture-shared design for RGB-X tracking, only activating the modality-specific parameters according to the input modality.
Besides, some unified methods~\cite{untrack, sutrack, xtrack, FlexTrack} have been proposed to learn the common latent space of any modality input, aiming at achieving more complete unification.
\textit{However, the overlook of distribution gap and the lack of multi-task inference capabilities remains incomplete for a powerful unified tracking and segmentation model.}

\textbf{Unified Task Methods.}
Meanwhile researchers have undertaken prominent efforts to unify tracking and segmentation tasks within specific modality. 
Existing unified task methods can be categorized into two main branches: the prompt based methods and the detection based methods.
The prompt based methods~\cite{utt,unicorn,omnivid} apply a shared appearance model for unified representation learning, and employ the delicately designed target prior to solve multiple subtasks.
Besides, the detection based methods~\cite{masa,uninext,omnitracker} introduce an external detectors to predict objects in various frames independently, and then establish the associations for tracking and segmentation.
\textit{However, these methods pay less attention to the multi-task gap, result in the degradation of task-specific knowledge during multi-task training.}

\section{Method}

\subsection{Overall Pipeline}
The proposed framework, termed SATA, comprises two core components: the Decoupled Mixture-of-Expert (DeMoE) mechanism and the Task-aware MOT (TaMOT) pipeline, as shown in Fig. \ref{fig3}. 
The DeMoE, combined with a powerful Transformer-based encoder, first yields the unified representation.  
Subsequently, the designed Task-aware MOT pipeline first generates tracking candidates based on task-specific prior knowledge, then maintains tracking of all candidates, thereby unifying all tracking and segmentation subtasks under the MOT paradigm.  

\subsection{Decoupled Mixture-of-Expert}
Given the input video frames of the RGB (R) modality and their corresponding auxiliary modalities (thermal, depth, or event modalities, collectively referred to as TDE), a weight-shared Transformer-based encoder~\cite{hivit} is employed to generate RGB tokens and TDE tokens.  
To enable the unified model to handle diverse input modalities, the proposed DeMoE is used to replace the feed-forward network (FFN) in each Transformer encoder layer, thereby converting different modality combinations into a unified token embedding format.  
DeMoE consists of three primary components: a Common-prompt Mixture of Expert (CpMoE), a Specific-activated Mixture of Expert (SaMoE), and Decoupling Learning.  

\begin{figure}[t]
\centering
\includegraphics[width=0.95\columnwidth]{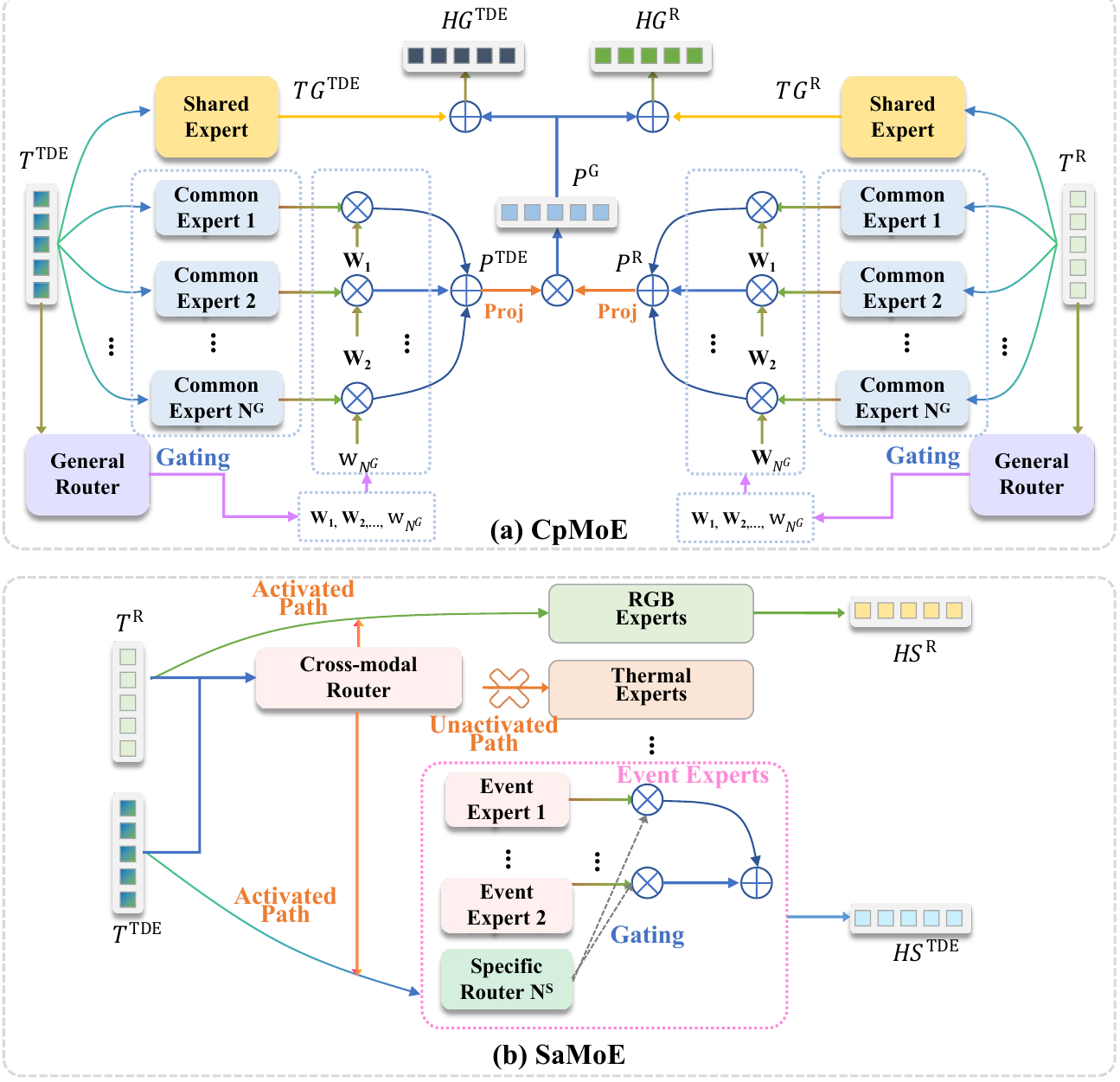}
\caption{Overview architecture of our CpMoE and SaMoE. (a) CpMoE.(b) SaMoE.}
\label{fig4}
\vspace{-0.5cm}
\end{figure}

\textit{Common-prompt Mixture of Expert.} 
At the $ l $-th Transformer encoder layer, our CpMoE takes the RGB token $ T^{\rm{R}}_l $ and TDE token $ T^{\rm{TDE}}_l $ from the output of the Multi-head Attention (MSA) block as input. It comprises a general router network $ \mathcal{R}^{\rm{G}} $, a shared expert $ g^{\rm{S}} $, and $ N^{\rm{G}} $ modality-common experts $\mathcal{G}^{\rm{G}}= \{g_{1}^{\rm{C}},...,g_{N^{\rm{G}}}^{\rm{C}}\} $.  
Specifically, as shown in Fig. \ref{fig4} (a), the shared expert $ g^{\rm{S}} $ directly copies weights from the corresponding FFN layer of the encoder and remains frozen during training to retain pre-trained general knowledge, yielding $ TG^{\rm{G}}_l $ and $ TG^{\rm{TDE}}_l $, i.e.,  
\begin{equation}
TG^{\rm{G}}_l = g^{\rm{S}}(T^{\rm{R}}_l), \quad TG^{\rm{TDE}}_l = g^{\rm{S}}(T^{\rm{TDE}}_l).   
\end{equation}  

Then, the modality-common experts $ g_{n}^{\rm{C}} $ ($ n = 1,...,N^{\rm{G}} $) are employed to generate modality-common prompts, transforming the general knowledge to be more suitable for the current input modalities. Each modality-common expert $ g_{n}^{\rm{C}} $ consists of two linear layers and a GELU activation layer.
The generated RGB prompt $P^{\rm{G}}_l$ and TDE prompt $P^{\rm{TDE}}_l$ are weighted sums of outputs from the top-K activated experts, i.e.,  
\begin{equation}
P^{\rm{R}}_l = \sum_{n=1}^{N^{\rm{G}}} p_n^{\rm{R}} g_{n}^{\rm{C}}(T^{\rm{R}}_l), \quad P^{\rm{TDE}}_l = \sum_{n=1}^{N^{\rm{G}}} p_n^{\rm{TDE}} g_{n}^{\rm{C}}(T^{\rm{TDE}}_l),
\end{equation}  
where $ p_n^{\rm{R}} $ and $ p_n^{\rm{TDE}} $ denote the gating values derived from the function $\mathcal{R}^{\rm{G}}$.

Subsequently, the modality-common prompt is obtained by performing element-wise multiplication on the RGB prompt and TDE prompt, which can be formulated as:  
\begin{equation}
P^{\rm{G}}_l = {\rm{proj}}(P^{\rm{R}}_l) \otimes {\rm{proj}}(P^{\rm{TDE}}_l),
\end{equation}  
where $ \otimes $ denotes the element-wise multiplication operation, and $ {\rm{proj}}(*) $ is the projection layer.  

Finally, the modality-common prompt is employed to map $ TG^{\rm{G}}_l $ and $ TG^{\rm{TDE}}_l $ into a cohesive token representation through element-wise addition, i.e.,  
\begin{equation}
HG^{\rm{R}}_l = P^{\rm{G}}_l \oplus TG^{\rm{R}}_l, \quad HG^{\rm{TDE}}_l = P^{\rm{G}}_l \oplus TG^{\rm{TDE}}_l,
\end{equation}  
where \( HG^{\rm{R}}_l \) and \( HG^{\rm{TDE}}_l \) refer to the prompted tokens for the RGB and TDE modalities, respectively, and \( \oplus \) denotes the element-wise addition operation.  

\textit{Specific-activated Mixture of Expert.} 
SaMoE is designed to capture modality-specific clues within multi-modal data, comprising intra-modal router networks ($\mathcal{R}^{\rm{R}}$ and $\mathcal{R}^{\rm{TDE}}$), a cross-modal router network $\mathcal{R}^{\rm{CM}}$, and several modality-specific experts ($\mathcal{G}^{X}$=$\{g_{1}^X,...,g_{N^{\rm{S}}}^X\} (X\rm{=R,TDE})$, where $N^{\rm{S}}$ denotes the number of modality-specific experts.  

As shown in Fig. \ref{fig4} (b), the cross-modal router network $\mathcal{R}^{\rm{CM}}$ activates the specific modality branch, thereby associating the current input modality with the corresponding modality experts. 
The modality-specific RGB tokens $HS^{\rm{R}}_l$ and TDE tokens $HS^{\rm{TDE}}_l$ are weighted sums of outputs from the top-K activated experts for each modality, i.e.,  
\begin{equation}
HS^{X}_l = \sum_{n=1}^{N^{\rm{S}}} s_n^X g_{n}^X(T^X_l), \quad X\rm{=R,TDE},
\end{equation}  
where $s_n^X$ represents the gating values derived from the function $\mathcal{R}^{X}$.

\textit{Unified Representation.} 
The RGB feature representation $F^{\rm{R}}_l$ and TDE feature representation $F^{\rm{TDE}}_l$ are obtained by aggregating these two types of embeddings from CpMoE and SaMoE with residual connections:  
\begin{equation}
F^X_l = HG^{X}_l \oplus HS^{X}_l \oplus T^{X}_l, \quad X\rm{=R,TDE}.
\end{equation}  
In the last layer ($L-th$ layer) of the encoder, the final unified feature representation $F^{\rm{U}}$ is obtained as:  
\begin{equation}
F^{\rm{U}} = F^{\rm{R}}_L \oplus F^{\rm{TDE}}_L.
\end{equation}  

\textit{Decoupling Learning.} 
To comprehensively model multi-modal representations, the learning processes of CpMoE and SaMoE adhere to two key principles: (a) promoting mutual complementarity among cross-modal experts; (b) avoiding information overlap between specific experts and common experts. Accordingly, we introduce two critical loss functions that act on the DeMoE.  

\underline{Cross-modal complementary learning}: To fully explore the complementary information across modalities, we randomly mask patches of one modality by assigning their values to a learnable token vector, generating masked features $\hat{HG}^{\rm{G}}_l$ and $\hat{HG}^{\rm{TDE}}_l$. The cross-modal complementary learning loss is then defined as:  
\begin{equation}
\resizebox{1.0\hsize}{!}{$
\mathcal{L}_{\text{CM
}} = \sum_{l=1}^{L} \text{MSE}(\hat{HG}^{\rm{R}}_l, {HG}^{\rm{R}}_l) + \sum_{l=1}^{L} \text{MSE}(\hat{HG}^{\rm{TDE}}_l, {HG}^{\rm{TDE}}_l),
$}
\end{equation}  
where $\text{MSE}(\cdot)$ denotes the mean squared error.  

\underline{Cross-expert orthogonal learning}: Additionally, we introduce an orthogonal loss to encourage independence between common and specific expert representations. This cross-expert orthogonal loss is defined as:  
\begin{equation}
\resizebox{1.0\hsize}{!}{$
\mathcal{L}_{\text{CE}} =  \sum_{l=1}^{L} \text{OPL}(\sum_{j=1}^{N^{\rm{S}}} \sum_{i=1}^{N^{\rm{G}}} g_i^{\rm{C}}(T^X_l), g_j^X(T^X_l)), X\rm{=R,TDE}.
$}
\end{equation}  
Here, $g_i^{\rm{C}}(*)$ and $g_j^X(*)$ represents the outputs of experts from $\mathcal{G}^{\rm{G}}$ and $\mathcal{G}^{X}$, respectively,
$\text{OPL}(*)$ denotes the orthogonal loss~\cite{orthogonal}. 

\subsection{Task-aware MOT pipeline}
To achieve the grand unification of tracking and segmentation while mitigating the representation gap across task domains, the proposed Task-aware MOT pipeline integrates all subtasks into the localization and association of multiple objects.  
Firstly, a Candidates Generation Module (CGM) is utilized to generate potential proposals based on task-specific prior information.  
Subsequently, the proposed Memory-enhanced Module (MEM) refines the historical features of each candidate, enabling efficient and effective modeling of temporal information.  
Finally, a simple bi-softmax nearest neighbor search is employed to achieve accurate matching between candidates and trajectories.  

\textit{Candidates Generation Module.}
Instead of introducing additional detectors to detect all potential targets in each frame~\cite{uninext,omnitracker}, we adopt a modified SAM2~\cite{sam2} as our foundation model to predict candidates based on the prior information of various tasks.  
Specifically, for SOT and VOS tasks, the initial target information (bounding box or mask) from the first frame can be used to generate the initial prompt token. In subsequent frames, masks predicted by the mask decoder with an affinity score exceeding a preset threshold are regarded as distractors; both the target and these distractors are defined as candidates.  
For MOT and MOTS tasks, we append a detection head to the foundation model to predict all potential objects in each frame. The box predictions are fed as multi-object prompts to the mask decoder, generating masks for all candidates.  
Once the positions of candidates in each frame are determined, we extract instance-level features by applying RoI Align.
Specifically, given the unified embedding $F^{\rm{U}}_t$ of $t-th$ frame and candidate coordinate $B^m_t$ corresponding to the $m-th$ tracklet, the corresponding candidate embedding can be acquired via $a^m_t = \mathrm{RoIAlign}(F^{\rm{U}}_t, B^m_t)$.

\begin{figure}[t]
\centering
\includegraphics[width=0.95\linewidth]{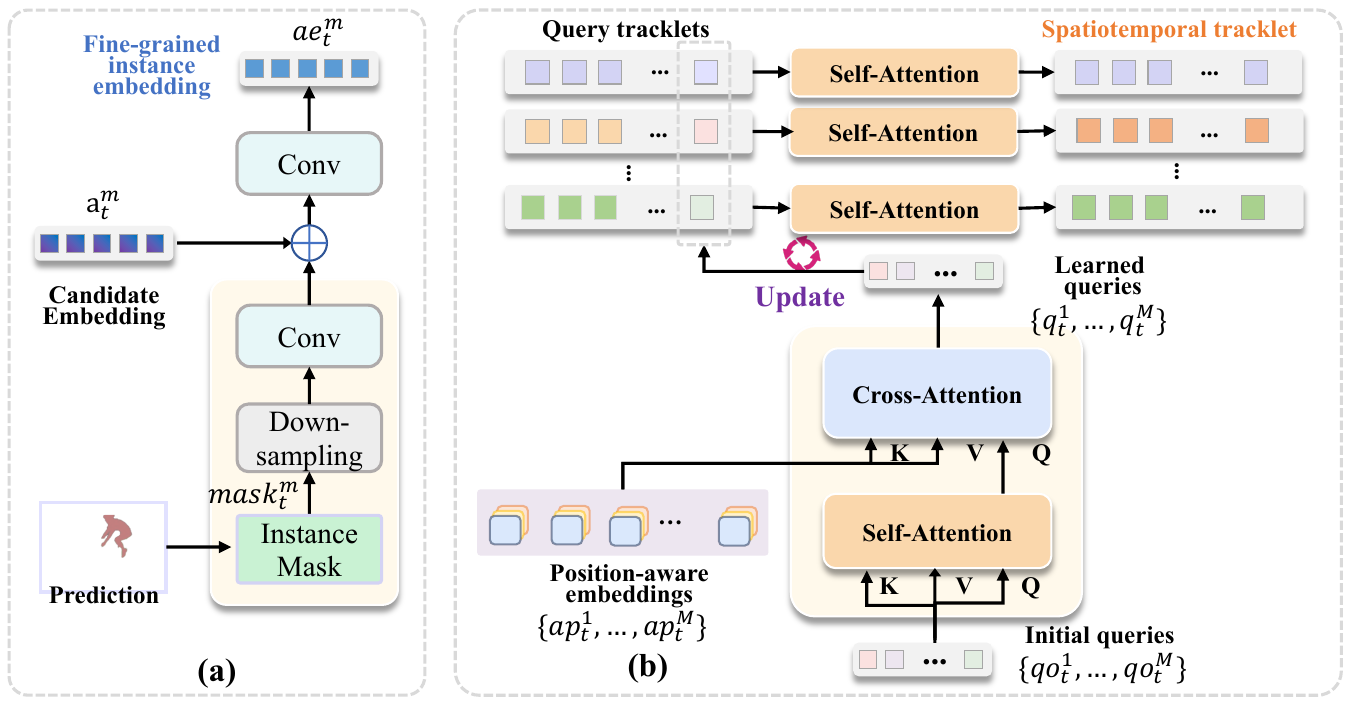}
\caption{Illustration of the fine-grained instance embeddings and spatiotemporal relationship modeling. (a) Fine-grained instance embeddings. (b) Spatiotemporal relationship modeling.}
\label{fig5}
\vspace{-0.5cm}
\end{figure}

\textit{Memory-enhanced Module.}
The memory-enhanced module comprises two components: (a) generating fine-grained instance embeddings; (b) effectively modeling spatiotemporal relationship, as shown in Fig. \ref{fig5}.

\underline{Fine-grained instance embeddings}: Candidate embeddings typically contain both foreground and background information at the edges, which leads to inaccurate instance matching when target objects cannot be clearly distinguished from the background.
To echo this problem, given the candidate embedding $a^m_t$, candidate coordinate $B^m_t$, and their corresponding mask prediction $mask^m_t$, the adjusted mask $\hat{mask}^m_t$ is first generated by down-sampling the predicted mask $mask^m_t$.
Then, the fine-grained instance embedding $ae^m_t$ is generated by:
\begin{equation}
ae^m_t = {\rm{ConV}}({\rm{ConV}}(\hat{mask}^m_t) \oplus a^m_t),
\end{equation}
where ${\rm{ConV}}(*, \theta_n)$ denotes the convolutional layer.

\begin{table*}[!http]
\begin{minipage}{0.375\linewidth}
\centering
\begin{subtable}{\linewidth}
\resizebox{1.0\linewidth}{!}{
\begin{tabular}{p{5.1cm}|p{0.65cm}p{0.65cm}p{0.68cm}|p{0.65cm}p{0.65cm}p{0.65cm}}
\hline
\small
\multirow{2}{*}{\textbf{\textcolor{green}{RGB SOT Method}}} & \multicolumn{3}{c|}{GOT10K} & \multicolumn{3}{c}{LaSOT} \\
\cline{2-7} &  AO  & SR$_{0.5}$ & SR$_{0.75}$ & AUC  & P$_{Norm}$ & P  \\ 
\hline 
\rowcolor{black!10} \multicolumn{7}{l}{Task-specific methods} \\ \hline  
GRM         ~\cite{grm}      & 73.4 & 82.9  &70.4 & 69.9 &79.3 & 75.8 \\ 
LORAT-G     ~\cite{lorat}    & 78.9 & 87.8  &80.7 & 76.2 &85.3 & 83.5 \\ 
ODTrack-B   ~\cite{odtrack}  & 77.0 & 87.9  &75.1 & 73.2 &83.2 & 80.6 \\ 
ARTrackV2-L ~\cite{artrackv2}& 77.5 & 86.0  &75.5 & 73.0 &82.0 & 79.6 \\ 
LMTrack     ~\cite{lmtrack}  & 80.1 & \textbf{91.5}  &79.0 & 73.2 &83.4 & 81.0 \\ 
TemTrack    ~\cite{temtrack} & 76.1 & 84.9  &74.4 & 73.1 &83.0 & 80.7 \\ 
MambaLCT    ~\cite{mambalct} & 76.2 & 86.7  &74.3 & 73.6 &84.1 & 81.6 \\ 
SPMTrack-L  ~\cite{spmtrack} & 80.0 & 89.4  &79.9 & 76.8 &\textbf{85.9} & 84.0 \\ 
MCITrack-B  ~\cite{mcitrack} & 77.9 & 88.2  &76.8 & 75.3 &85.6 & 83.3\\ 
DreamTrack-L~\cite{dreamtrack}&79.9 & 88.5  &80.3 & 76.6 &85.6 & 83.1 \\ 
SAM2.1++    ~\cite{sam2++}   & 81.1 & -     &-    & 75.1 &-    &-     \\ 
\rowcolor{black!10} \multicolumn{7}{l}{Unified modality methods} \\ \hline
\rowcolor{cyan!5} SUtrack-L   ~\cite{sutrack}  & 81.5 & 89.5  &83.3 & 75.2 &84.9 &83.2  \\
\hline 
\rowcolor{black!10} \multicolumn{7}{l}{Unified task methods} \\ \hline
\rowcolor{green!5} UTT         \cite{utt}      & 67.2 & 76.3  &60.5 & 64.6 &-    &67.2  \\ 
\rowcolor{green!5} Unicorn     \cite{unicorn}  & -    &-      & -   & 68.5 &76.6 &74.1  \\ 
\rowcolor{green!5} UNINEXT-H   \cite{uninext}  & -    &-      & -   & 72.2 &80.7 &79.4  \\ 
\rowcolor{green!5} OmniViD     \cite{omnivid}  & -    &-      & -   & 70.8 &79.6 &76.9  \\
\rowcolor{green!5} SAM2.1      \cite{sam2}     & 80.7 & -     &-    & 70.5 &-    &-     \\ 
\rowcolor{green!5} OmniTracker-L\cite{omnitracker}& - &-      & -   & 69.1 &77.3 &75.4  \\ 
\hline
\rowcolor{purple!5} SATA     & \textbf{81.3} & 91.4  &\textbf{83.7} & \textbf{77.3} &85.7 &\textbf{84.6}  \\
\bottomrule
\end{tabular}}
\caption{SOTA comparisons on RGB SOT.}
\label{Tab1}
\end{subtable}
\end{minipage}
\hfill
\begin{minipage}{0.306\linewidth}
\centering
\begin{subtable}{\linewidth}
\resizebox{1.0\linewidth}{!}{
\begin{tabular}{p{5.1cm}|p{0.65cm}p{0.65cm}|p{0.65cm}p{0.65cm}}
\hline
\multirow{2}{*}{\textbf{\textcolor{orange}{RGB-T SOT Method}}} & \multicolumn{2}{c|}{LasHeR} & \multicolumn{2}{c}{RGBT234} \\
\cline{2-5}  & PR    & SR  & PR    & SR    \\ 
\hline 
\rowcolor{black!10} \multicolumn{5}{l}{Modality-specific methods} \\ \hline 
TBSI+~\cite{tbsi+}           &75.5 &59.6  &91.0 &67.0\\
TransMMSTC~\cite{siammmstc}  &72.3 &57.4  &88.6 &65.7\\ 
TransTIH~\cite{siamtih}      &72.0 &57.2  &89.4 &66.4\\
PURA ~\cite{pura}            &-    &-     &93.3 &70.3\\ 
AINet~\cite{ainet}           &74.2 &59.1  &89.2 &67.3\\ 
CAFormer~\cite{caformer}     &70.0 &55.6  &88.3 &66.4\\
AETrack~\cite{aetrack}       &74.7 &59.6  &91.6 &68.8\\ 
DMD~\cite{dmd}               &72.6 &57.6  &89.3 &66.7\\ 
\hline 
\rowcolor{black!10} \multicolumn{5}{l}{Architecture-shared methods} \\ \hline 
\rowcolor{blue!5} ViPT~\cite{vipt}             &65.1 &52.5  &83.5 &61.7\\
\rowcolor{blue!5} SDSTrack~\cite{sdstrack}     &66.5 &53.1  &84.8 &62.5\\
\rowcolor{blue!5} OneTrack~\cite{onetracker}   &67.2 &53.8  &85.7 &64.2\\
\rowcolor{blue!5} GMMT~\cite{gmmt}             &70.7 &56.6  &87.9 &64.7\\
\rowcolor{blue!5} CSTrack~\cite{cstrack}       &75.6 &60.8  &94.0 &70.9\\
\rowcolor{blue!5} STTrack~\cite{sttrack}       &76.0 &60.3  &89.8 &66.7\\
\hline
\rowcolor{black!10} \multicolumn{5}{l}{Unified modality methods} \\ \hline 
\rowcolor{cyan!5} UnTrack~\cite{untrack}       &-    &-     &84.2 &62.5\\
\rowcolor{cyan!5} FlexTrack~\cite{FlexTrack}   &77.3 &-     &92.7 &69.9\\
\rowcolor{cyan!5} XTrack~\cite{xtrack}         &73.1 &58.7  &87.8 &65.4\\
\rowcolor{cyan!5} SUTrack-L~\cite{sutrack}     &76.9 &-     &93.7 &70.3\\
\hline
\rowcolor{purple!5} SATA      &\textbf{77.8} &\textbf{61.7}  &\textbf{94.3} &\textbf{71.5} \\
\bottomrule
\end{tabular}}
\caption{SOTA comparisons on RGB-T SOT.}
\label{Tab2}
\end{subtable}
\end{minipage}
\hfill
\begin{minipage}{0.306\linewidth}
\centering
\begin{subtable}{\linewidth}
\resizebox{\linewidth}{!}{
\begin{tabular}{p{5.1cm}|p{0.65cm}p{0.65cm}|p{0.65cm}p{0.65cm}}
\hline
\multirow{2}{*}{\textbf{\textcolor{blue}{RGB-D SOT Method}}} & \multicolumn{2}{c|}{DepthTrack} & \multicolumn{2}{c}{VOTRGBD} \\ 
\cline{2-5}                 & PR  & Re     & EAO & Acc \\ \hline
\rowcolor{blue!5} STTrack~\cite{sttrack}      &63.2 &63.4   &-    &-    \\
\rowcolor{blue!5} CSTrack~\cite{cstrack}      &65.2 &66.4   &77.4 &83.3 \\
\rowcolor{blue!5} OneTrack~\cite{onetracker}  &60.7 &60.4   &72.7 &81.9 \\
\hline
\rowcolor{cyan!5} UnTrack~\cite{untrack}      &61.3 &61.0   &72.1 &81.5 \\
\rowcolor{cyan!5} FlexTrack~\cite{FlexTrack}  &67.1 &66.9   &78.0 &83.8\\
\rowcolor{cyan!5} XTrack-L~\cite{xtrack}      &65.4 &64.3   &74.0 &82.8\\
\rowcolor{cyan!5} SUTrack-L~\cite{sutrack}    &66.5 &66.4   &76.6 &83.5 \\
\hline
\rowcolor{purple!5} SATA     &\textbf{67.9} &\textbf{67.6}   &\textbf{78.4} &\textbf{84.1}  \\
\hline 
\end{tabular}}
\caption{SOTA comparisons on RGB-D SOT.}
\vspace{-0.1cm}
\label{Tab3}
\end{subtable}
\begin{subtable}{\linewidth}
\resizebox{\linewidth}{!}{
\begin{tabular}{p{5.1cm}|p{0.65cm}p{0.65cm}|p{0.65cm}p{0.65cm}}
\hline
\multirow{2}{*}{\textbf{\textcolor{pink}{RGB-E SOT Method}}} & \multicolumn{2}{c|}{VisEvent} & \multicolumn{2}{c}{COESOT} \\ 
\cline{2-5}                & PR  &AUC   & PR  & SR     \\ \hline
CSAM~\cite{csam}           &81.6 &-     &76.7 &68.3\\
\hline
\rowcolor{blue!5} STTrack~\cite{sttrack}     &78.6 &61.9   &-    &-    \\
\rowcolor{blue!5} CSTrack~\cite{cstrack}     &82.4 &65.2   &77.4 &83.3 \\
\rowcolor{blue!5} \textbf{}SDSTrack~\cite{sdstrack}   &76.7 &59.7   &-    &-   \\
\rowcolor{blue!5} OneTrack~\cite{onetracker} &76.7 &60.8   &-    &-   \\
\hline
\rowcolor{cyan!5} FlexTrack~\cite{FlexTrack} &81.4 &64.1   &-    &-    \\
\rowcolor{cyan!5} XTrack-L~\cite{xtrack}     &80.5 &63.3   &-    &-    \\
\rowcolor{cyan!5} SUTrack-L~\cite{sutrack}   &80.5 &63.8   &-    &-    \\
\rowcolor{cyan!5} UnTrack~\cite{untrack}     &76.3 &58.7   &-    &-   \\
\hline
\rowcolor{purple!5} SATA    &\textbf{82.8} &\textbf{66.7}   &\textbf{80.4} &\textbf{71.6}\\
\hline 
\end{tabular}}
\caption{SOTA comparisons on RGB-E SOT.}
\label{Tab4}
\end{subtable}
\end{minipage}
\caption{SOTA comparisons on RGB, RGB-Depth, RGB-Thermal, and RGB-Event SOT tasks.}
\vspace{-0.2cm}
\end{table*}

\begin{table*}[!http]
\begin{minipage}{0.5\linewidth}
\centering
\begin{subtable}{\linewidth}
\resizebox{1.0\linewidth}{!}{
\begin{tabular}{p{5.1cm}|p{0.9cm}p{0.9cm}p{0.9cm}|p{0.9cm}p{0.9cm}p{0.9cm}}
\hline
\small
\multirow{2}{*}{\textbf{\textcolor{green}{RGB MOT Method}}} & \multicolumn{3}{c|}{BDD} & \multicolumn{3}{c}{DanceTrack} \\
\cline{2-7} & IDF1 & mMOTA & MOTA & IDF1  &HOTA  & MOTA\\ 
\hline 
\rowcolor{black!10} \multicolumn{7}{l}{Task-specific methods} \\ \hline 
DiffMOT~\cite{diffmot}      &-     &-     &-    &63.0 &62.3 &\textbf{92.8} \\ 
Hybrid-SORT~\cite{hybrid-sort}&-   &-     &-    &67.4 &65.7 &91.8 \\ 
\rowcolor{black!10} \multicolumn{7}{l}{Unified task methods} \\ \hline 
\rowcolor{green!5} Unicorn ~\cite{unicorn}     & 71.3 & 41.2 & 66.6 &-    &-    &-    \\
\rowcolor{green!5} UNINEXT-H ~\cite{unicorn}   & 69.9 & 44.2 & 67.1 &-    &-    &-    \\
\rowcolor{green!5} MASA-SAM-H~\cite{masa}      & 71.7 & 44.5 & -    &-    &-    &-    \\
\rowcolor{green!5} SAM2MOT-Co~\cite{sam2mot}   & 70.8 & -    & 57.5 &83.4 &75.5 &89.2 \\
\hline
\rowcolor{purple!5} SATA     &\textbf{73.2}  &\textbf{46.3} &\textbf{67.8} &\textbf{83.7}  &\textbf{76.1} &90.5\\ \hline
\end{tabular}}
\caption{SOTA comparisons on RGB MOT}
\label{Tab5}
\end{subtable}
\end{minipage}
\hfill
\begin{minipage}{0.42\linewidth}
\centering
\begin{subtable}{\linewidth}
\resizebox{1.0\linewidth}{!}{
\begin{tabular}{l|c|ccccc}
\hline
{\textbf{\textcolor{orange}{RGB-T MOT Method}}} & HOTA  & DetA & AssA & MOTA & IDF1  \\ 
\hline
\rowcolor{black!10} \multicolumn{6}{l}{Task-specific methods} \\ \hline 
Bytetrack~\cite{bytetrack} &53.4 &58.2 &49.6 &72.4 &64.9  \\ 
MOTRv2 ~\cite{motrv2}      &53.1 &51.2 &\textbf{55.8} &63.3 &64.4  \\ 
HGT-Track~\cite{hgt-track} &54.0 &61.3 &48.1 &71.1 &60.9  \\ 
\rowcolor{black!10} \multicolumn{6}{l}{Unified task methods} \\ \hline 
\rowcolor{green!5} Unitrack~\cite{unitrack}   &46.9 &46.1 &48.4 &57.1 &57.4  \\ 
\rowcolor{green!5} Unicorn~\cite{unicorn}     &49.4 &48.3 &51.2 &59.0 &60.6  \\ 
\rowcolor{green!5} UnisMOT~\cite{unirtl}      &54.2 &59.5 &49.7 &65.7 &60.3  \\ 
\hline
\rowcolor{purple!5} SATA    &\textbf{59.7} &\textbf{63.6} &53.7 &\textbf{75.1} &\textbf{65.4}  \\ 
\bottomrule
\end{tabular}}
\caption{SOTA comparisons on RGB-T MOT}
\label{Tab6}
\end{subtable}
\end{minipage}
\caption{SOTA comparisons on RGB, and RGB-Thermal MOT tasks.}
\vspace{-0.4cm}
\end{table*}

\underline{Spatiotemporal relationship modeling}: We first employ several MLP layers on the normalized coordinate $\hat{B}^m_t$ to generate the positional embedding $p^m_t$.
The position-aware embedding can be represented as $ap^m_t = p^m_t \oplus a^m_t$.
Then, we introduce the learnable queries $q^m_t$ to capture video spatiotemporal information via the same architecture as the Querying Transformer (Q-Former)~\cite{blip2}.
Specifically, the initial queries $\{qo^1_{t}, qo^2_{t},...,qo^M_{t}\}$ are fed into the self-attention layer to model spatial interactions.  
After that, the interacted queries and the position-aware embeddings $\{ap^1_{t}, ap^2_{t},...,ap^M_{t}\}$ are fed into the cross-attention layer, obtaining the $t-th$ learned query $\{q^1_{t}, q^2_{t},...,q^M_{t}\}$.
Finally, we construct the temporal correlations on each query tracklet $\{q^m_{t-1}, q^m_{t-2},...,q^m_{t-T}\} (m=1,...,M)$ via a single self-attention layer, generating the spatiotemporal tracklet $\{qe^m_{t-1}, qe^m_{t-2},...,qe^m_{t-T}\}$.

Finally, the comprehensive representation $f^M_t$ of the $M-th$ instances in $t-th$ frame can be transformer by concatenating the fine-grained instance embedding $ae^M_t$ and the learned query $q^M_t$, and the comprehensive tracklet can be represented by $\Gamma^n=\{f^n_{t-1}, f^n_{t-2},...,f^n_{t-T}\} (n=1,...,N)$.

\textbf{Instance matching.}
Given the instance $f^m_t$ and the tracklet $\Gamma^n$ generated by the candidate generation module and the memory-enhanced module, the similarity computation can be represented by:
\begin{equation}
\begin{array}{cc}
s(f^m_t, \Gamma^n)=\frac{1}{2}(s_1(f^m_t, \Gamma^n)+s_2(f^m_t, \Gamma^n)), \\
s_1(f^m_t, \Gamma^n)=\frac{1}{2}\left[\frac{\exp(f^m_t\cdot\Gamma^n)}{\sum^M_{m=1}
\exp(f^m_t\cdot\Gamma^n)}+\frac{\exp(f^m_t\cdot\Gamma^n)}{\sum_{n=1}^N 
\exp(f^m_t\cdot\Gamma^n)}\right], \\
s_2(f^m_t, \Gamma^n)=\frac{f^m_t\cdot}{\|f^m_t\|\|\Gamma^n\|}. 
\end{array} 
\end{equation}
With the above similarity function, we can obtain the final assignment matrix $A\in \mathbb{R}^{N \times M}$. Finally, we match the $\epsilon-th$ tracklet with $m-th$ instances via $\text{argmax}(s(f^m_t, \Gamma^1),...,s(f^m_t, \Gamma^N))=s(f^m_t, \Gamma^\epsilon)$.

\section{Experiments}

\subsection{Implementation Details}
The SATA model is developed using Python 3.8 and PyTorch 1.11. The training process leverages 8 NVIDIA A100 GPUs, while the inference speed is evaluated on a single NVIDIA 3090TI GPU.

\textit{Architecture.} Our model uses HiViT-L~\cite{hivit} as the transformer encoder, and we select SAM2~\cite{sam2} as our foundation model in the Task-aware MOT pipeline. 
The transformer encoder and the foundation model are initialized with the pre-trained parameters of SAM2~\cite{sam2}.

\textit{Training.}   
During the stage of candidate generation, the affinity mask scores, object predictions, and IoU predictions of the mask decoder are optimized by MAE loss, cross-entropy loss, and $L_1$ loss, respectively~\cite{sam2}. 
In addition, the decoupled MoE loss $L_{\text{MoE}}=\mu L_{\text{CM}}+\lambda L_{\text{CE}}$ in the DeMoE is employed to promote comprehensive learning of the unified embedding.
In the stage of instance matching, the partial supervision loss and self-supervised loss in KeepTrack~\cite{keeptrack} are employed to supervise the assignment matrix $A$ for SOT and VOS datasets, and the cross-entropy loss is applied to optimize SATA for MOT and MOTS datasets.

\subsection{State-of-the-Art Comparisons}

We compare SATA with state-of-the-art methods on 18 large-scale benchmarks with 4 types of input (i.e., RGB, RGB-T, RGB-D, RGB-E) and 4 subtasks (i.e., SOT, VOS, MOT and MOTS).

\begin{table*}[htbp]
\begin{minipage}{0.365\linewidth}
\centering
\begin{subtable}{\linewidth}
\resizebox{1.0\linewidth}{!}{
\begin{tabular}{p{4.9cm}|p{0.65cm}p{0.65cm}p{0.65cm}|p{0.65cm}p{0.65cm}p{0.65cm}}
\hline
\multirow{2}{*}{\textbf{\textcolor{green}{RGB VOS Method}}}  & \multicolumn{3}{c|}{DAVIS 2016 val} & \multicolumn{3}{c}{DAVIS 2017 val} \\
\cline{2-7}   &  $\mathcal{J}\&\mathcal{F}$ & $\mathcal{J}$ & $\mathcal{F}$ & $\mathcal{J}\&\mathcal{F}$ & $\mathcal{J}$ & $\mathcal{F}$  \\ \hline
\rowcolor{black!10} \multicolumn{7}{l}{Task-specific methods} \\ \hline
XMem~\cite{xmem}        &92.0 &90.7 &93.2 &87.7 &84.0 &91.4  \\
DEVOS~\cite{devos}      &92.8 &\textbf{91.8} &93.8 &88.2 &84.5 &91.9  \\
OneVOS~\cite{onevos}    &92.7 &91.0 &94.3 &88.5 &84.6 &92.4  \\
M3-VOS~\cite{m3vos}     &-    &-    &-    &-    &\textbf{86.0} &-     \\
SAM2~\cite{sam2}                &-    &-    &-    &88.9 &-    &-     \\
TAM-S~\cite{sam2}              &-    &-    &-    &89.2 &-    &-     \\
\midrule
\rowcolor{black!10} \multicolumn{7}{l}{Unified task methods} \\ \hline
\rowcolor{green!5} UniTrack~\cite{unicorn}   &-    &-    &-    &-    &58.4 &-     \\
\rowcolor{green!5} Unicorn-T~\cite{unicorn}]  &83.2 &83.0 &83.4 &64.5 &62.7 &66.3  \\
\rowcolor{green!5} Unicorn-ConvL~\cite{unicorn}  &87.4 &86.5 &88.2 &69.2 &65.2 &73.2  \\
\rowcolor{green!5} UNINEXT-R50~\cite{uninext}     &-    &-    &-    &74.5 &71.3 &77.6  \\
\rowcolor{green!5} UNINEXT-L~\cite{uninext}       &-    &-    &-    &77.2 &73.2 &81.2  \\
\rowcolor{green!5} UNINEXT-H~\cite{uninext}       &-    &-    &-    &81.8 &77.7 &85.8  \\
\rowcolor{green!5} OmniTracker-T~\cite{omnitracker}  &84.7 &84.1 &85.3 &66.2 &64.9 &67.5  \\
\rowcolor{green!5} OmniTracker-L~\cite{omnitracker}   &88.5 &87.3 &89.7 &71.0 &66.8 &75.2  \\
\midrule
\rowcolor{purple!5} SATA      &\textbf{93.4} &\textbf{91.6} &\textbf{95.2} &\textbf{89.7} &\textbf{86.1} &\textbf{93.0}    \\
\bottomrule
\end{tabular}}
\caption{Comparisons on RGB VOS.}
\label{Tab7}
\end{subtable}
\end{minipage}
\hfill
\begin{minipage}{0.355\linewidth}
\centering
\begin{subtable}{\linewidth}
\resizebox{1.0\linewidth}{!}{
\begin{tabular}{p{4.6cm}|p{0.65cm}p{0.65cm}p{0.65cm}|p{0.65cm}p{0.65cm}p{0.65cm}}
\hline
\multirow{2}{*}{\textbf{\textcolor{orange}{RGB-T VOS Method}}}  & \multicolumn{3}{c|}{VisT300} & \multicolumn{3}{c}{VTUAV} \\
\cline{2-7}   &  $\mathcal{J}\&\mathcal{F}$ & $\mathcal{J}$ & $\mathcal{F}$ & $\mathcal{J}\&\mathcal{F}$ & $\mathcal{J}$ & $\mathcal{F}$  \\ \hline
\rowcolor{black!10} \multicolumn{7}{l}{Modality-specific methods} \\ \hline
STM~\cite{stm}     &60.4 &57.9 &62.8 &-    &-    &-      \\
STCN~\cite{stcn}   &71.4 &74.4 &73.8 &65.5 &61.0 &69.9  \\
STCN-T~\cite{stcn} &72.3 &-    &-    &-    &-    &-  \\
TBD~\cite{TBD}          &70.5 &68.3 &72.6 &-    &-    &-     \\
CFBI+~\cite{CFBI}       &74.1 &71.8 &76.4 &-    &-    &-     \\
D3S~\cite{d3s}          &-    &-    &-    &57.0 &53.4 &60.7  \\
AlpahRefine~\cite{alpharefine}&-    &-    &-    &65.9 &59.9 &71.9  \\
AOT~\cite{aot}          &76.8 &74.0 &79.6 &-    &-    &-     \\
AOT-B~\cite{aot}        &-    &-    &-    &81.8 &77.7 &85.8  \\
AOT-L~\cite{aot}        &-    &-    &-    &82.0 &77.5 &86.5  \\
XMem~\cite{xmem}        &75.7 &73.3 &78.0 &69.1 &65.1 &73.1  \\
XMem-T~\cite{xmem}      &77.9 &-    &-    &-    &-    &-     \\
ViTNet~\cite{vist300}   &81.9 &79.2 &84.5 &76.7 &72.9 &80.8  \\
\midrule
\rowcolor{black!10} \multicolumn{7}{l}{Architecture-shared methods} \\ \hline
\rowcolor{blue!5} X-Prompt~\cite{xprompt} &84.2 &81.7 &86.7 &87.3 &82.8 &91.8  \\
\midrule
\rowcolor{purple!5} SATA &\textbf{87.4} &\textbf{84.5} &\textbf{90.3} &\textbf{88.5} &\textbf{84.4}\textbf{} &\textbf{92.6}  \\
\bottomrule
\end{tabular}}
\caption{Comparisons on RGB-T VOS.}
\label{Tab8}
\end{subtable}
\end{minipage}
\hfill
\begin{minipage}{0.25\linewidth}
\centering
\begin{subtable}{\linewidth}
\resizebox{\linewidth}{!}{
\begin{tabular}{p{4.6cm}|p{0.65cm}p{0.65cm}p{0.65cm}}
\hline
\multirow{2}{*}{\textbf{\textcolor{blue}{RGB-D VOS Method}}}  & \multicolumn{3}{c}{ARKitTrack} \\
\cline{2-4}   &  $\mathcal{J}\&\mathcal{F}$ & $\mathcal{J}$ & $\mathcal{F}$ \\ \hline
STCN-D~\cite{stcn}          &53.7 &49.8 &57.5  \\
ARKitVOS~\cite{arkittrack}  &66.2 &62.5 &69.8  \\
XMem~\cite{xmem}            &71.6 &68.5 &74.6   \\
DeAOT~\cite{deaot}          &72.6 &70.0 &75.3  \\
AOT-L-Swin~\cite{aot}       &77.8 &75.0 &80.7  \\
\rowcolor{blue!5} X-Prompt~\cite{xprompt}     &82.1 &79.4 &84.9  \\
\midrule
\rowcolor{purple!5} SATA     &\textbf{85.1} &\textbf{82.8} &\textbf{87.4}   \\
\bottomrule
\end{tabular}}
\caption{Comparisons on RGB-D VOS.}
\label{Tab9}
\end{subtable}
\begin{subtable}{\linewidth}
\resizebox{\linewidth}{!}{
\begin{tabular}{p{4.6cm}|p{0.65cm}p{0.65cm}p{0.65cm}}
\hline
\multirow{2}{*}{\textbf{\textcolor{pink}{RGB-E VOS Method}}}  & \multicolumn{3}{c}{LLE-VOS} \\
\cline{2-4}   &  $\mathcal{J}\&\mathcal{F}$ & $\mathcal{J}$ & $\mathcal{F}$ \\ \hline
STCN~\cite{stcn}            &47.4 &45.0 &49.8  \\
XMem~\cite{xmem}            &54.2 &59.0 &49.4  \\
AOT-B~\cite{aot}            &62.3 &64.9 &59.6  \\
DeAOT~\cite{deaot}          &63.3 &65.3 &61.4  \\
LLE-VOS~\cite{llevos}       &67.8 &70.2 &65.4  \\
\midrule
\rowcolor{purple!5} SATA     &\textbf{71.4} &\textbf{73.6} &\textbf{69.1}    \\
\bottomrule
\end{tabular}}
\caption{Comparisons on RGB-E VOS.}
\label{Tab10}
\end{subtable}
\end{minipage}
\vspace{-0.2cm}
\caption{SOTA comparisons on RGB, RGB-Depth, RGB-Thermal, and RGB-Event VOS tasks.}
\vspace{-0.4cm}
\end{table*}

\textbf{SOT.} 
The results of RGB SOT are presented in Tab.~\ref{Tab1}. Our model achieves 81.3\% AO and 77.3\% AUC on GOT10K~\cite{got10k} and LaSOT~\cite{lasot}, respectively, surpassing the recent RGB tracker SAM2.1++~\cite{sam2++}, SUTrack~\cite{sutrack}, and LMTrack~\cite{lmtrack} by 0.2\%/2.2\%, 0.3\%/2.1\%, and 1.2\%/4.1\%, respectively.
In addition, compared with existing unified task models, e.g., UTT~\cite{utt}, Unicorn~\cite{unicorn}, and OmniTracker~\cite{omnitracker}, SATA achieves performance gains of 12.7\%, 8.8\%, and 7.9\% in AUC on LaSOT, respectively. 
Besides, SATA sets a new state-of-the-art on 6 multi-modal SOT benchmarks, as illustrated in Tab.~\ref{Tab2}, Tab.~\ref{Tab3}, and Tab.~\ref{Tab4}.
On LasHeR~\cite{lasher}, DepthTrack~\cite{depthtrack}, and VisEvent~\cite{visevent}, SATA surpassing the recent best unified modality trackers, XTrack~\cite{xtrack}, SUTrack~\cite{sutrack} and FlexTrack~\cite{FlexTrack}, by 4.7\%/2.5\%/2.3\%, 0.9\%/1.4\%/2.3\%, and 0.5\%/0.8\%/1.4\% in PR, respectively.

\textbf{MOT.} 
We report the results of RGB MOT in Tab.~\ref{Tab5}. Our SATA achieves the best MOTA, HOTA, and IDF1 scores on DanceTrack~\cite{danceTrack} and BDD~\cite{bdd100k}. Specifically, it outperforms all previous unified task methods, achieving an mMOTA score of 67.8\% on BDD, surpassing Unicorn~\cite{unicorn}, UNINEXT-H~\cite{uninext} and SAM2MOT-Co~\cite{sam2mot} by 1.2\%, 0.7\%, and 10.3\%, respectively.
In addition, as shown in Tab.~\ref{Tab6}, SATA achieves state-of-the-art performance on UniRTL~\cite{unirtl} with RGB-T input, outperforming the previous best method UnisMOT~\cite{unirtl} by 5.3\% in HOTA. 

\textbf{VOS.} 
We present a comparison with recent advanced RGB VOS methods in Table \ref{Tab7}, reporting accuracy using standard protocols. SATA shows significant improvement over the best existing methods, achieving the highest scores of 93.4\% and 89.7\% $\mathcal{J}\&\mathcal{F}$ on these DAVIS datasets~\cite{davis2017}.
Besides, as presented in Table \ref{Tab8}, Table \ref{Tab9}, and Table \ref{Tab10}, SATA secures the top positions on the RGB-T~\cite{vist300,vtuav}, RGB-D~\cite{arkittrack}, and RGB-E VOS~\cite{llevos} benchmarks. 

\textbf{MOTS.} 
We evaluate SATA's MOTS capability on BDD MOTS~\cite {bdd100k}
As shown in Table~\ref{Tab11}, our approach outperforms existing advanced unified models, i.e.,  Unicorn~\cite{unicorn} and UNINEXT-H~\cite{uninext} by noteworthy margins of 8.5\% and 2.4\% in mMOTSA, respectively.

\begin{table}
\centering
\resizebox{0.95\linewidth}{!}{
\begin{tabular}{l|cccc}
\hline
\textbf{\textcolor{green}{RGB MOTS Method}} &mMOTSA &mMOTSP &mIDF1 &ID Sw\\ 
\hline
\rowcolor{black!10} \multicolumn{5}{l}{Task-specific methods} \\ \hline
MaskTrackRCNN~\cite{MaskTrackRCNN}  &12.3 &59.9 &26.2 &9116 \\ 
STEm-Seg~\cite{STEm-Seg}            &12.2 &58.2 &25.4 &8732 \\
QDTrack-mots~\cite{QDTrack-mots}    &22.5 &59.6 &40.8 &1340 \\
PCAN~\cite{pcan}                    &27.4 &66.7 &45.1 &876  \\
VMT~\cite{vmt}                      &28.7 &67.3 &45.7 &825  \\
MASA-B~\cite{masa}                  &35.2 &-    &49.2 &-    \\
MASA-H~\cite{masa}                  &35.8 &-    &49.7 &-    \\
\hline
\rowcolor{black!10} \multicolumn{5}{l}{Unified task methods} \\ \hline
\rowcolor{green!5} Unicorn~\cite{unicorn}    &29.6 &67.7 &44.2 &1731 \\
\rowcolor{green!5} UNINEXT-L~\cite{uninext}  &32.0 &60.2 &45.4 &1634 \\
\rowcolor{green!5} UNINEXT-H~\cite{uninext}  &35.7 &68.1 &48.5 &1776 \\
\hline
\rowcolor{purple!5} SATA   &\textbf{38.1} &\textbf{72.3} &\textbf{52.4} &\textbf{721}  \\ 
\bottomrule
\end{tabular}
}
\caption{SOTA comparisons on RGB MOTS tasks.}
\vspace{-0.5cm}
\label{Tab11}
\end{table}

\begin{table}[!http]
    
    \centering
\resizebox{0.95\linewidth}{!}{
\begin{tabular}{cl|ccccc}
\toprule
\multirow{3}{*}{Type} & \multirow{3}{*}{Method} & \multicolumn{1}{c}{\underline{SOT}} & 
\multicolumn{1}{c}{\underline{SOT}} & 
\multicolumn{1}{c}{\underline{VOS}} & 
\multicolumn{1}{c}{\underline{MOT}} & 
\multicolumn{1}{c}{\underline{MOTS}} \\
& &GOT10K&LasHeR&LLE-VOS&UniRTL&BDD MOTS  \\
& &(AO)&(PR)&($\mathcal{J\&F}$)&HOTA&MOTSA\\ 
\midrule 
\rowcolor{purple!5} & SATA & \textbf{81.3}&\textbf{77.8}&\textbf{71.4}&\textbf{59.7}&\textbf{38.1}  \\ 
\midrule
\multirow{4}{*}{DeMoE} 
& W/o CpMoE &80.8&75.8&70.7&56.2&37.9  \\
& W/o SaMoE &81.3&75.3&69.3&55.3&37.7  \\ 
& W/o CpMoE \& SaMoE &80.8 &74.7&67.9&56.1&37.7  \\
& W/o $L_{\text{MoE}}$ &80.7&75.2&70.2&58.4&36.9  \\  
                          \midrule
\multirow{4}{*}{TaMOT} 
&W/o CGM             &78.5&74.3&68.7&-   &-  \\
&W/o fine-grained memory   &80.7&75.3&69.2&56.7&34.1  \\
&W/o spatiotemporal memory &79.7&75.8&67.4&57.5&30.7 \\  
&W/o MEM                   &79.5&74.3&67.0&54.2&29.1 \\ 
\midrule
\end{tabular}}
\caption{Ablation studies on DeMoE and TaMOT.}
\vspace{-0.5cm}
\label{Tab12}
\end{table}

\subsection{Ablation Studies}
In this section, we conduct component-wise analysis for a better understanding of our method. 
The methods are evaluated on 5 benchmarks (GOT10K~\cite{got10k}, LasHeR~\cite{lasher}, LLE-VOS~\cite{llevos}, UniRTL~\cite{unirtl}, and BDD MOTS~\cite{bdd100k}) from 4 subtasks (SOT, VOS, MOT, MOTS).

\textbf{Decoupled Mixture-of-Expert.}
To investigate the impact of our proposed DeMoE, several versions of our proposed method are provided, including 
\ding{172}: Removing the CpMoE sub-module in DeMoE. 
\ding{173}: Removing the SaMoE sub-module in DeMoE. 
\ding{174}: Removing the CpMoE and SaMoE sub-modules in DeMoE. 
\ding{175}: Removing the MoE loss $L_{\text{MoE}}$ in DeMoE. 
As can be seen in Table \ref{Tab12}, the performance degrades significantly after removing CpMoE or SaMoE sub-modules in multi-modal subtasks, which demonstrates the effectiveness of the proposed DeMoE in handling any modality input.

\textbf{Task-aware MOT pipeline.}
To further verify the effectiveness of the proposed TaMOT, several variants are designed, including 
\ding{172}: Removing the CGM. 
\ding{173}: Removing the fine-grained memory.
\ding{174}: Removing the spatiotemporal memory. 
\ding{175}: Removing the MEM.
As shown in Table \ref{Tab12}, comparative results indicate that keeping track of all potential objects can further improve the robustness of SOT and VOS tasks. Besides, the tracking performance experiences a significant decline upon the ablation of fine-grained memory or spatiotemporal memory, which confirms the necessity of the comprehensive trajectory information incorporated in TaMOT for accurate instance matching.

\section{Conclusion}
In this paper, we propose a universal tracking and segmentation framework, referred to as SATA, which is capable of processing inputs from any modality and predicting results for a wide range of tracking and segmentation subtasks with a fully shared network architecture, model weights, and inference pipeline.
Extensive experiments on 18 challenging benchmarks demonstrate that SATA achieves superior performance across these tasks. We hope that SATA can lay a solid foundation for future research on AGI.

\bigskip

\section{Acknowledgments}
This work was supported in part by National Natural Science Foundation of China under Grant No.62441235, and part by the Beijing Natural Science Foundation under Grant No.L257005.

\small
\bibliography{aaai2026}

\section*{Appendix}

In this supplementary material, we first elaborate on the detailed specifications of the model. Subsequently, we outline the training procedure of the proposed SATA framework. Following this, we present more details about the loss functions. Finally, we present more experimental results along with further in-depth analyses of our SATA approach.

\section{Model details}
Here, we further elaborate on the architectural details of the proposed SATA, encompassing the Backbone Network, Decoupled Mixture-of-Experts, Candidates Generation Module, Memory-enhanced Module, and Instance Matching.

\subsection{Backbone Network}
We selected Hiera-L~\cite{hiera} as our backbone network, which comprises four stages for both low-level and high-level visual modeling, as presented in Table~\ref{tab1}. Each stage consists of multiple Hiera blocks, which are analogous to the typical Transformer block. Each hiera block incorporates normalization layers, a multi-head attention block, a feed-forward network, and several residual connections. To fully leverage the open-world general knowledge acquired from large-scale training data, we employ the pretrained weights from SAM2~\cite{sam2} for parameter initialization.

\begin{table}[!http]
\centering
\resizebox{1.0\linewidth}{!}{
\begin{tabular}{c|ccc|c}
\toprule
Model   &Channels            &Blocks     &Heads       &Param \\ \hline
Hiera-T &[96-192-384-768]    &[1-2-7-2]  &[1-2-4-8]   &28M  \\
Hiera-B+&[112-224-448-896]   &[2-3-16-3] &[2-4-8-16]  &70M  \\
Hiera-L &[144-288-576-1152]  &[2-6-36-4] &[2-4-8-16]  &214M \\
\bottomrule
\end{tabular}}
\caption{Configuration for Hiera variants. Channels, Blocks and Heads specify the channel width, number of Hiera blocks and heads in each block for the four stages, respectively. The stride of the four stages are [4, 8, 16, 32].
Our SATA employ the Hiera-L~\cite{hiera} as our backbone network.
}
\label{tab1}
\end{table}

\subsection{Decoupled Mixture-of-Expert}

DeMoE consists of two primary components: a Common-prompt Mixture of Expert (CpMoE) and Specific-activated Mixture of Expert (SaMoE).

\textit{CpMoE.} 
Each modality-common expert $ g_{n}^C $ consists of two linear layers and a GELU activation layer: the first linear layer projects the input to a lower dimension, and the second layer projects it back to the original dimension.  
In the experiments, we project each input token with channel $c$ to the significantly lower
dimensional latent space $k$ ($k = c/8$).
Each CpMoE contains 4 modality-common experts, and we activate the top-2 experts during inference.

\textit{SaMoE.} 
In the designed SaMoE, the modality-specific experts shared the same architecture as that in the modality-common experts.
SaMoE contains 4 modality-specific experts in each modality, and we also activate the top-2 experts in each modality during inference.

\begin{figure*}[!t]
\centering
\includegraphics[width=0.6\linewidth]{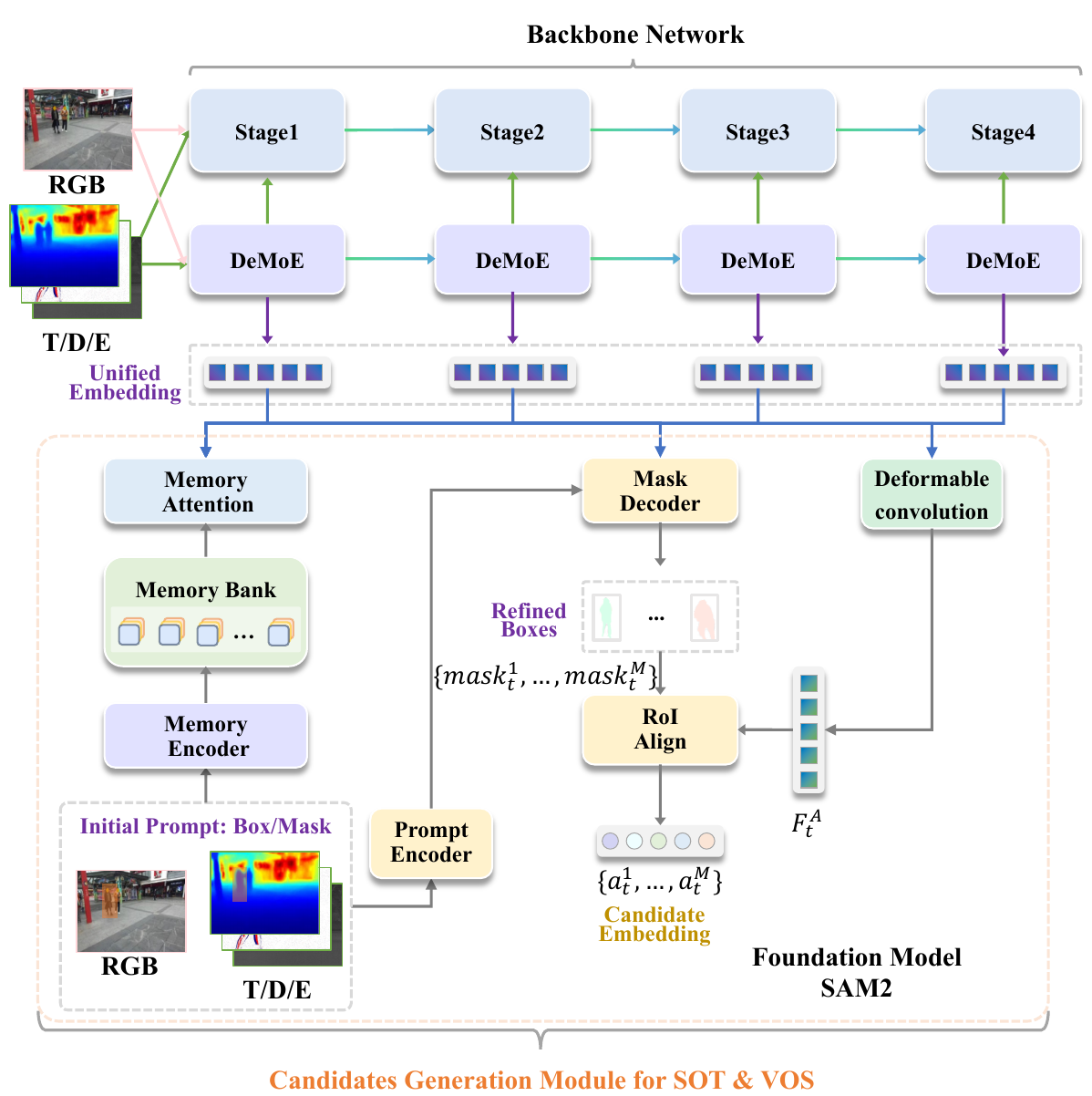}
\caption{Details of our Candidates Generation Module (CGM) for SOT\&VOS.}
\label{fig1}
\end{figure*}

\subsection{Candidates Generation Module}

The designed Candidates Generation Module adopt a modified SAM2~\cite{sam2++} our foundation model to predict candidates based on the prior information of various tasks.
In the following content, we first briefly revisit the employed foundation model-SAM2.
Then, we introduce the process of generating potential candidates for SOT\&VOS and MOT\&MOTS, respectively.

\textit{Revisiting SAM2.}
SAM2 consists of four main components: (i) image encoder, (ii) prompt encoder, (iii) memory bank, and (iv) mask decoder. 

\underline{Image encoder} applies ViT Hiera backbone to embed the input image. Our SATA uses the unified embedding generated by the proposed DeMoE.

\underline{The prompt encoder} takes two types of prompts, including sparse (e.g., points, bounding boxes) and dense (e.g., masks). The prompt tokens output by the prompt encoder can be represented as $x_{prompt}$. 

\underline{The mask decoder} is designed to take the memory-conditioned image embeddings produced by the memory attention layer along with the prompt tokens from the prompt encoder as its inputs. It can then generate a set of predicted masks $\{mask_{1},...mask_{N}\}$, along with the corresponding mask affinity score $\{s_{mask,1},...s_{mask,N}\}$, and a single occlusion score $s_{occ}$ for each frame.
The final output of the mask decoder is selected based on the highest affinity score.

\underline{The memory bank} consists of the encoded initialization frame with a user-provided segmentation mask and six recent frames with segmentation masks generated by the tracking output.
After the mask decoder generates output masks, the output mask is passed through a memory encoder to obtain a memory embedding. A new memory is created after each frame is processed.
Please refer \cite{sam2} for more details.

\textit{Candidates Generation for SOT\&VOS.}
For SOT and VOS tasks, initial target information is used to generate the initial prompt token and predict candidates with affinity scores, as shown in Fig. \ref{fig1}. Given the predicted masks $\{\text{mask}_1, \ldots, \text{mask}_{N'}\}$ and their corresponding mask affinity scores $\{s_{\text{mask},1}, \ldots, s_{\text{mask},N'}\}$, we select candidates only when $s_{\text{occ}} > 0$ and $s_{\text{mask}}$ exceeds the preset threshold $\tau_{\text{mask}}$. We then generate $M$ candidates per frame. In our experiments, $\tau_{\text{mask}}$ is set to 0.7. The masks $\{\text{mask}_t^1, \ldots, \text{mask}_t^M\}$ are used to generate bounding boxes $\{B_t^1, \ldots, B_t^M\}$.

\textit{Candidates Generation for MOT\&MOTS.}
For MOT and MOTS tasks, we append an additional detection head to the foundation model to predict all potential objects in each frame, as shown in Fig. \ref{fig2}.

\underline{Detection head.}
Given unified embeddings generated by DeMoE with scale ratios of $\frac{1}{4}, \frac{1}{8}, \frac{1}{16}, \frac{1}{32}$ (from Stage 1 to Stage 4 of the backbone network), we first use deformable convolution~\cite{masa} to dynamically fuse multi-scale features, producing unified image embeddings $F_t^A$ for the $t$-th frame. We then employ 4 stacked blocks (incorporating task-aware and scale-aware attention from Dynamic Head~\cite{masa}), a classification predictor, and a box regressor to predict all potential instances.

\underline{Boxes refinement.}
As illustrated in our manuscript, predicted boxes from the detection head are fed as multi-object prompts to the mask decoder, generating masks $\{\text{mask}_t^1, \ldots, \text{mask}_t^M\}$ for all candidates in the $t$-th frame. Leveraging SAM2's strong capability in predicting fine-grained instance masks, we use these masks to correct the bounding boxes generated by the additional detection head, yielding refined boxes $\{B_t^1, \ldots, B_t^M\}$.

\underline{Instance embedding.}
Using the refined boxes $\{B_t^1, \ldots, B_t^M\}$, we extract instance embeddings $\{a_t^1, \ldots, a_t^M\}$ via RoI-Align applied to the unified image embeddings $F_t^A$.

\begin{figure*}[!t]
\centering
\includegraphics[width=0.6\linewidth]{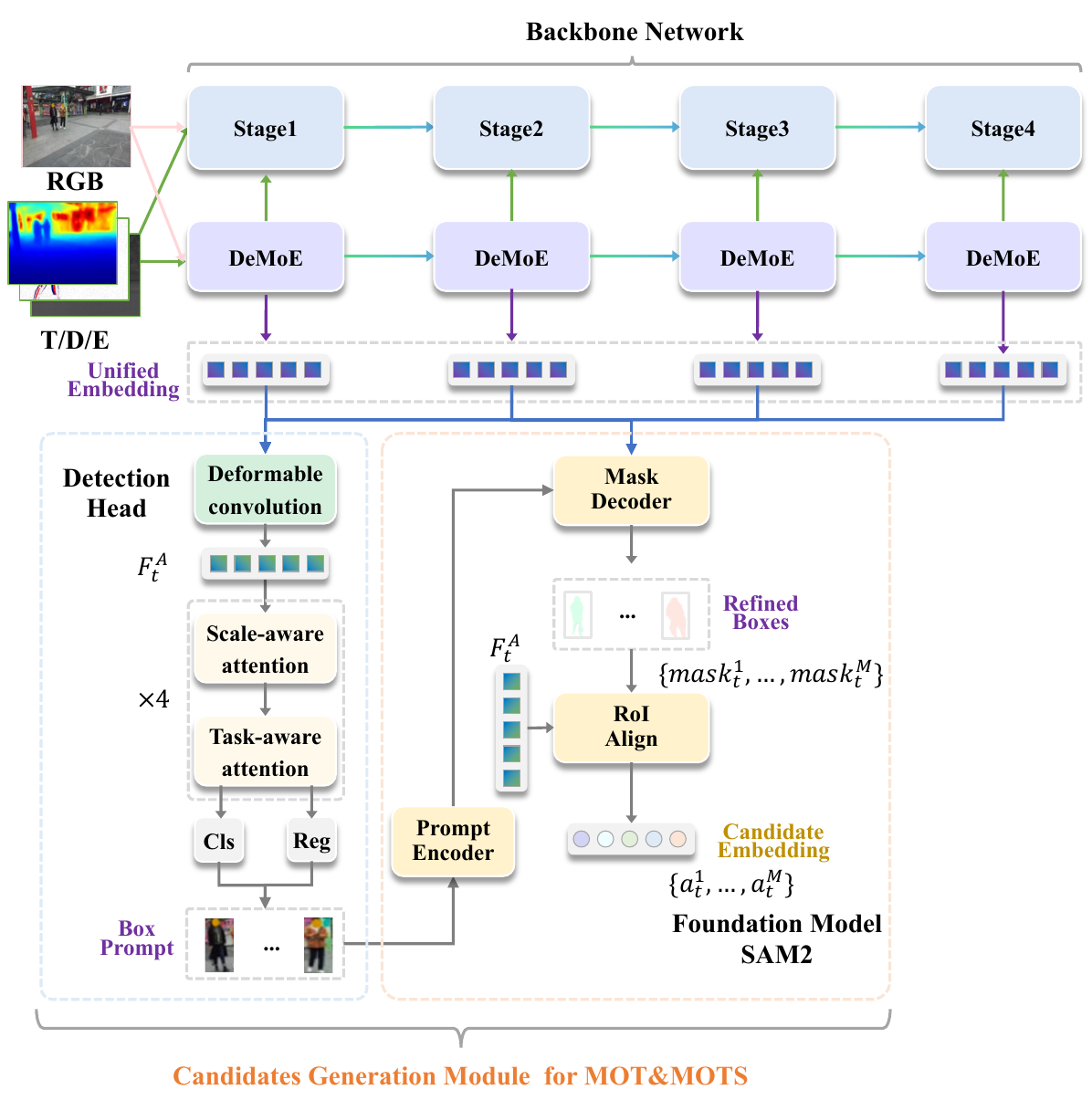}
\caption{Details of our Candidates Generation Module (CGM) for MOT\&MOTS.}
\label{fig2}
\end{figure*}

\subsection{Memory-enhanced Module}

\textit{Instance queries.}
In the proposed MEM module, we introduce learnable queries $q_t^m \in \mathbb{R}^{D \times C}$ ($m=1,\ldots,M$) for each candidate to capture video spatiotemporal information, employing the same architecture as the Querying Transformer (Q-Former)~\cite{blip2}. Here, $D$ denotes the number of learned queries, and $C$ represents the channel dimension. In our experiments, the Q-Former outputs 8 tokens for each instance to enable more efficient storage of historical information.

\begin{figure*}[!http]
\centering
\includegraphics[width=0.85\linewidth]{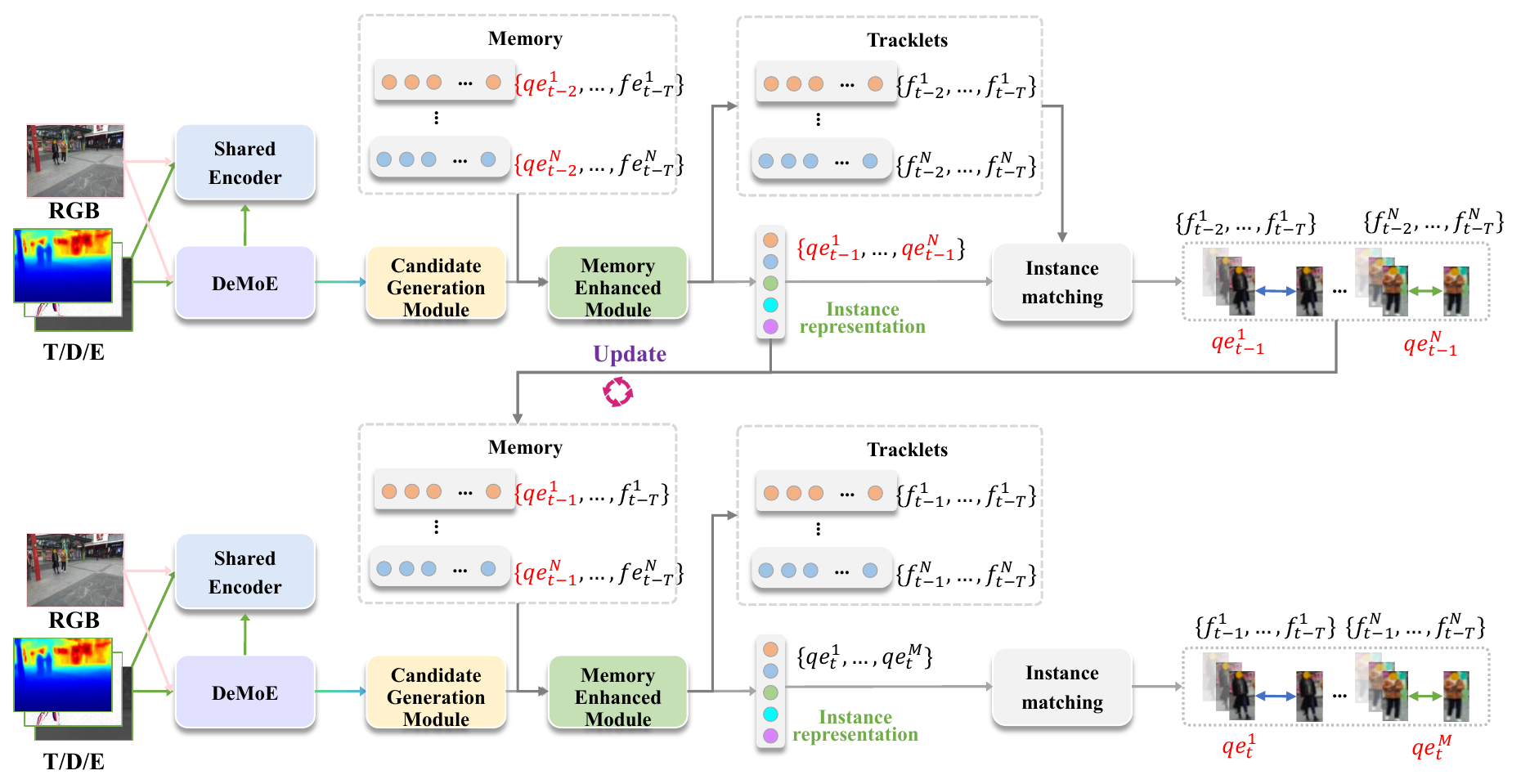}
\caption{Illustration of memory updating pipeline.}
\label{fig3}
\end{figure*}

\begin{table*}[t]
\centering
\resizebox{1.0\linewidth}{!}{
\begin{tabular}{c|c|cc|cccc}
\toprule
Stage & Task & Dataset & Sampling Weight & Batch Size & Num GPU & Lr  & Step \\
\midrule
\multirow{2}{*} {\uppercase\expandafter{\romannumeral1}}& \multirow{1}{*} {RGB OD} & COCO~\cite{coco} & 0.9 &  \multirow{2}{*}{16} & \multirow{2}{*}{8} & \multirow{2}{*}{$2e^{-4}$} & \multirow{2}{*}{180K}  \\
& \multirow{1}{*} {RGB-T OD} & UniTRL~\cite{unirtl} & 0.1 &   &  &  &   \\
\midrule
\multirow{16}{*} {\uppercase\expandafter{\romannumeral2}}& \multirow{4}{*} {RGB SOT} & LaSOT~\cite{coco} & 0.2 &  \multirow{16}{*}{32} & \multirow{16}{*}{8} & \multirow{16}{*}{$3e^{-4}$} & \multirow{16}{*}{360K}  \\
& & GOT10K~\cite{got10k} & 0.2 &   &  &  &   \\
& & TrackingNet~\cite{trackingnet} & 0.2 &   &  &  &   \\
& & COCO~\cite{coco} & 0.2 &   &  &  &   \\
\cline{2-4}
& RGB-T SOT& LasHeR~\cite{lasher} & 0.33  &   &  &  &  \\
& RGB-D SOT& DepthTrack~\cite{depthtrack} & 0.33 &   &  &  &   \\
& RGB-E SOT& VisEvent~\cite{visevent} & 0.33  &   &  &  &  \\
\cline{2-4}
& \multirow{3}{*} {RGB VOS} & DAVIS~\cite{davis2017} & 0.33 &   &  &  &   \\
& & YouTube~\cite{youtube} & 0.33 &   &  &  &   \\
& & MOSE~\cite{mose} & 0.33 &   &  &  &   \\
\cline{2-4}
& RGB-T VOS& VisT300~\cite{visevent} & 0.25  &   &  &  &  \\
& RGB-D VOS& ArKitTrack~\cite{arkittrack} & 0.25 &   &  &  &   \\
& RGB-D VOS& DepthTrack~\cite{depthtrack} & 0.25 &   &  &  &   \\
& RGB-E VOS& LLE-VOS~\cite{llevos} & 0.25  &   &  &  &  \\
\cline{2-4}
& RGB MOT\&MOTS & BDD~\cite{bdd100k} & 0.8  &   &  &  &  \\
& RGB-T MOT & UniTRL~\cite{unirtl} & 0.2  &   &  &  &  \\
\bottomrule
\end{tabular}}
\caption{Details in training. 'OD' denotes object detection.} 
\label{tab2}
\end{table*}

\textit{Memory updating.}
Assuming that the Candidates Generation Module (CGM) generates $M$ candidates for the $t$-th frame, while the Memory-enhanced Module stores $N$ tracklets $\Gamma^n = \{f^n_{t-1}, f^n_{t-2}, \ldots, f^n_{t-T}\}$ ($n=1,\ldots,N$). Candidates that fail to match any tracklet are assigned new IDs, and tracklets that do not match any instances over $T^E$ consecutive frames are terminated, where $T^E$ is experimentally set to 50. Historical tracklets are updated only after candidates are matched with them, and spatiotemporal relationships are calculated to facilitate instance matching in the next frame. The default length $T$ of historical tracklets is set to 15. The memory and its management mechanism are visualized in Fig. \ref{fig3}.

\subsection{Instance matching}
This sub-section details the algorithm for instance matching association. Our proposed SATA employs a simple bi-softmax nearest neighbor search to generate the affinity matrix $A$. To avoid ambiguous candidate-tracklet matching, only pairs with affinity exceeding $\tau_{th}=0.75$ can be associated. Subsequently, the Hungarian algorithm~\cite{keeptrack} is applied to generate the tracking output.

\section{Training Process}
The overall training process consists of two stages: Detection Training and Mixture Training. We use the AdamW optimizer with a weight decay of 0.05. Detailed training hyperparameters are provided in Table \ref{tab2}. The model is trained on 8 A100 GPUs with a base learning rate of $2e^{-4}$. A StepLR learning rate scheduler is employed in each stage, where the learning rate is reduced by a factor of 10 after specified steps. Multi-scale training is adopted: original images are resized such that the shortest side ranges between 480 and 800 pixels, while the longest side does not exceed 1333 pixels~\cite{omnitracker}.

\subsection{Detection Training}
In the first stage, we initialize with the official SAM2 weights~\cite{sam2} and keep all SAM2 components frozen during training. Only the parameters of the additional detection head in the CGM are trainable. The model is trained on the COCO~\cite{coco} and UniTRL~\cite{unirtl} datasets with a batch size of 16 for 180K iterations.

\subsection{Mixture Training}
In the second stage, parameters of DeMoE, CGM, and MEM are set to be trainable. Our training data includes commonly used datasets for SOT and VOS tasks: COCO~\cite{coco}, LaSOT~\cite{lasot}, GOT-10k~\cite{got10k}, TrackingNet~\cite{trackingnet}, DepthTrack~\cite{depthtrack}, VisEvent~\cite{visevent}, LasHeR~\cite{lasher}, DAVIS~\cite{davis2017}, YouTube~\cite{youtube}, MOSE~\cite{mose}, VisT300~\cite{vist300}, ARKitTrack~\cite{arkittrack}, and LLE-VOS~\cite{llevos}; as well as datasets for MOT and MOTS tasks: BDD~\cite{bdd100k} and UniRTL~\cite{unirtl}. Training is conducted with a batch size of 32 for 360K iterations. For SOT\&VOS, each batch contains mixed data sampled from these datasets, with RGB data sampled at twice the rate of multi-modal data, following the approach in SUTrack~\cite{sutrack}. To ensure balanced performance across various benchmarks, we set the data sampling ratios as (SOT\&VOS):(MOT\&MOTS) = 0.6:0.4.

\section{Loss functions}
First, the original loss function $L_\text{ORI}$ employed in SAM2 are used across all stages.
Please find more details in SAM2~\cite{sam2}.
Secondly, to promote the unified representation learning, we employ the MoE loss $L_\text{MoE}$ proposed in DeMoE.
Thirdly, to associate instances on different frames, since only the targets’ locations are provided in the existing SOT\&VOS dataset, the partial supervision loss and self-supervised loss in KeepTrack\cite{keeptrack} are employed to supervise the affinity matrix $A$ generated by TaMOT.
For MOT\&MOTS tasks, the cross-entropy loss is applied to optimize SATA.

\begin{figure*}[!http]
\centering
\includegraphics[width=0.85\linewidth]{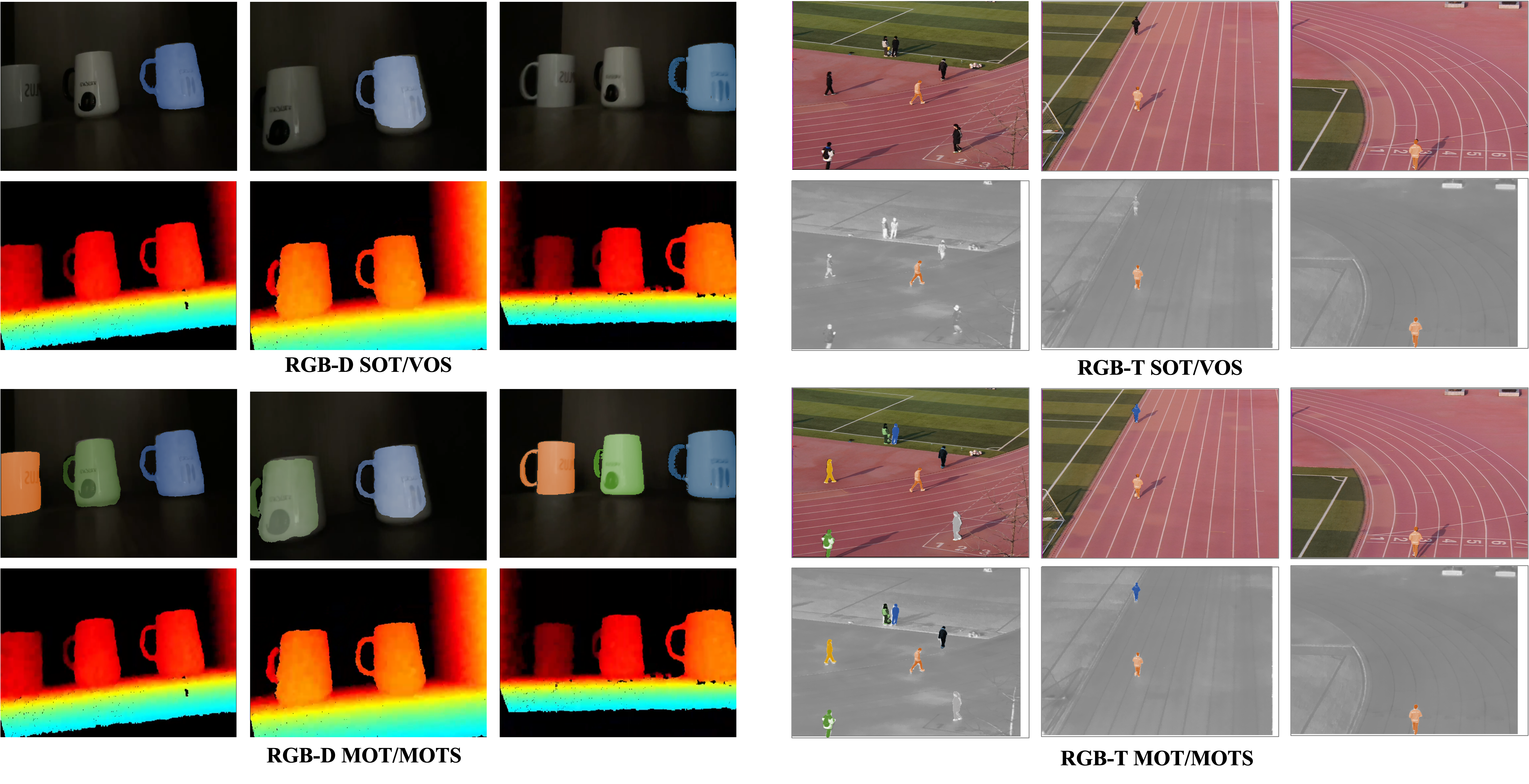}
\caption{Visualizations of tracking results predicted by SATA.}
\label{fig4}
\end{figure*}
 
\subsection{MoE supervision}

\textit{Cross-modal complementary learning loss.} In order to promote the mutual complementarity among cross-modal experts, as illustrated in our manuscript, the cross-modal complementary learning loss is defined as:  
\begin{equation}
\resizebox{1.0\hsize}{!}{$
\mathcal{L}_{\text{CM
}} = \sum_{l=1}^{L} \text{MSE}(\hat{HG}^{R}_l, {HG}^{R}_l) + \sum_{l=1}^{L} \text{MSE}(\hat{HG}^{TDE}_l, {HG}^{TDE}_l),
$}
\end{equation}  
where $\text{MSE}(\cdot)$ denotes the mean squared error.  

\textit{Cross-expert orthogonal loss.} 
Furthermore, there exists a risk that multiple experts may converge to similar functions, which could undermine the overall effectiveness of Mixture-of-Expert. To address this issue, we design a cross-expert orthogonal loss that encourages each expert to specialize in distinct aspects of the data. This loss enhances the ability of experts to learn diverse information from multi-modal data and is defined as follows:
\begin{equation}
\resizebox{1.0\hsize}{!}{$
\mathcal{L}_{\text{CE}} =  \sum_{l=1}^{L} \text{OPL}(\sum_{j=1}^{N^S} \sum_{i=1}^{N^G} g_i^C(T^X_l), g_j^X(T^X_l)), X=R,T,D,E
$}
\end{equation}  
Here, $g_i^C(*)$ and $g_j^X(*)$ represents the outputs of experts from $\mathcal{G}^{G}$ and $\mathcal{G}^{X}$, respectively,
$\text{OPL}(*)$ denotes the orthogonal loss~\cite{orthogonal}, which can be defined as:
\begin{equation}
\begin{array}{cc}
     \text{OHL} = \sum_{j=1}^{N^G} \sum_{\substack{k=1 \\ k \neq j}}^{N^S} \frac{\langle \tilde{x}_{j}, \tilde{x}_{k} \rangle}{\langle \tilde{x}_{k}, \tilde{x}_{k} \rangle} \tilde{x}_{k}, 
\end{array}
\end{equation}
where $\langle*\rangle$ denotes the inner product between two vector, $\tilde{x}_{j}$ denotes the output of expert $g_j^C(*)$ for input $x$, $\tilde{x}_{k}$ denotes the output of expert $g_k^X(*)$ for input $x$.

\textit{Modality-aware loss.} 
In addition, to supervise the cross-modal router and ensure accurate prediction of the input modality, we use the predicted probabilities to compute a cross-entropy loss against the true task labels \( y^{task} \):
\begin{equation}
L_{\text{TASK}} = \sum_{n=1}^{L} \sum_{j=1}^{K} \left( -y_{\text{task}}(j) \cdot \log(s^{CM}_n(j)) \right),
\end{equation}
where \( K \) denotes the number of tasks, \( y_{\text{task}}(j) \) represents the ground-truth label for task \( j \), and \( s^{CM}_n(j) \) denotes the predicted probability for task \( j \).

\subsection{SOT/VOS supervision}
In existing SOT/VOS datasets, only target objects and their corresponding locations are provided for training. To effectively supervise the learning of TaMOT, we employ both the partial supervision loss and self-supervision loss from KeepTrack~\cite{keeptrack}.

\textit{Partially Supervised Loss.} We formulate the problem of associating instances across two consecutive frames as obtaining an affinity matrix \( A \) between the two candidate sets. For candidate \( v^{t}_i \) in frame \( t \) that corresponds to candidate \( v^{t-1}_j \) in frame \( t-1 \), we set \( A_{ij}=1 \); otherwise, \( A_{ij}=0 \). For each consecutive frame pair in a video sequence, we retrieve the single candidate corresponding to the annotated target. 
For the candidates \( \{v^{t-T}_i, \ldots, v^{t}_i\} \), the assignment matrix \( A \) generated by our proposed TaMOT reflects the association between \( v^{t}_i \) and \( v^{t-1}_i \). The supervised loss is then defined as the negative log-likelihood of the assignment probability:
\begin{equation}
L_{\text{sup}} = -\log \boldsymbol{A}_{i,i}.
\end{equation}

\textit{Self-Supervised Loss.} To enhance the robustness of TaMOT, given a predicted candidate set \( \mathcal{V}^{t} \) and its corresponding ground-truth association set \( \mathcal{C} = \{(i,i)\}_{i=1}^N \), we generate a series of candidate sets \( \{\mathcal{V}^{t-T}, \ldots, \mathcal{V}^{t-1}\} \) from \( \mathcal{V}^{t} \) via feature augmentation. This augmentation involves randomly translating candidate locations, randomly adjusting candidate affinity scores, and transforming the input image before extracting features. The self-supervised loss is formulated as:
\begin{equation}
L_{\text{self}} = \sum_{(i,j) \in \mathcal{C}} -\log \boldsymbol{A}_{i,j}.
\end{equation}

Finally, we combine the self-supervised loss and partially supervised loss as \( L_{\text{TaMOT-S}} = L_{\text{self}} + L_{\text{sup}} \).

\begin{figure}[!http]
\centering
\includegraphics[width=1.0\linewidth]{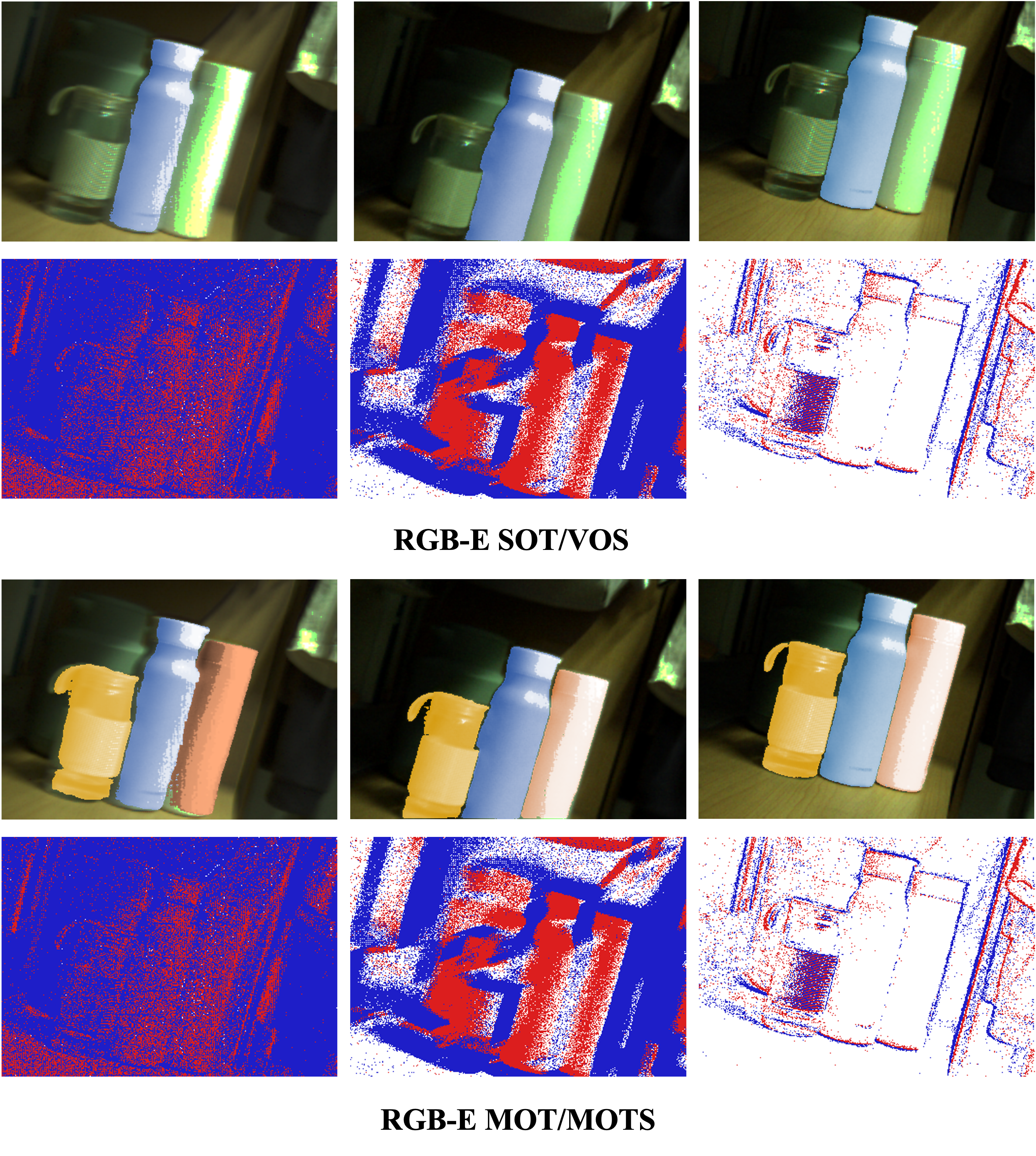}
\caption{Visualizations of tracking results predicted by SATA.}
\label{fig5}
\end{figure}

\subsection{MOT supervision}
For MOT/MOTS tasks, given the ground-truth assignment matrix \( A^g \), we formulate the prediction of the assignment matrix \( A \) as a binary classification problem. Accordingly, we employ cross-entropy loss to optimize the network:
\begin{equation}
\resizebox{1.0\hsize}{!}{$
\mathcal{L}_{\text{TaMOT-M}} = \frac{-1}{MN} \sum_{i=1}^N \sum_{j=1}^M \left[ A_{i,j}^g \log(A_{i,j}) + (1 - A_{i,j}^g) \log(1 - A_{i,j}) \right],
$}
\end{equation}
where \( A_{i,j} \) and \( A_{i,j}^g \) denote the \((i,j)\)-th elements of the predicted assignment matrix \( \boldsymbol{A} \) and the ground-truth assignment matrix \( A^g \), respectively.

\section{Additional Experiments}

\subsection{Visualizations}
As illustrated in Fig. \ref{fig4} and Fig. \ref{fig5}, we visualize the tracking results of SATA across different tasks. SATA can accurately locate targets based on their state in the initial frame. Additionally, with the designed TaMOT pipeline, our SATA is capable of tracking multiple objects belonging to specific categories. By consistently tracking multiple potential objects, SATA effectively alleviates task conflicts and bridges the data distribution gap between various downstream tasks. Notably, even in the absence of RGB-D and RGB-E training data for MOT\&MOTS tasks, SATA can still precisely predict object positions and establish accurate associations.

\subsection{Ablation studies}

\textit{Unified Embedding.}
To investigate the effectiveness of the proposed DeMoE, we implement several variants by replacing DeMoE with existing methods, including:
\ding{172}: Replacing DeMoE with the Shared Embedding from UnTrack~\cite{untrack}.
\ding{173}: Replacing DeMoE with the MeMoE from XTrack~\cite{xtrack}.
\ding{174}: Replacing DeMoE with HMOE-Fuse from FlexTrack~\cite{FlexTrack}.
\ding{175}: Replacing DeMoE with the Unified Modality Representation from SUTrack~\cite{sutrack}.

We conduct ablation studies on RGB-X SOT tasks, evaluating all models on LasHeR~\cite{lasher} (PR score), DepthTrack~\cite{depthtrack} (PR score), and VisEvent~\cite{visevent} (PR score). As shown in Table \ref{tab3}, the proposed DeMoE outperforms all these variants.

\begin{table}[!http]
    \centering
    \resizebox{0.95\linewidth}{!}{
    \begin{tabular}{c|ccc}
\toprule
\multirow{3}{*}{Method} & \multicolumn{1}{c}{\underline{RGB-T SOT}} & 
\multicolumn{1}{c}{\underline{RGB-D SOT}} & 
\multicolumn{1}{c}{\underline{RGB-E SOT}} \\
Method   & LasHeR & DepthTrack &  VisEvent  \\ 
&(PR)&(PR)&(PR)\\ 
\hline
\rowcolor{purple!5} SATA & 77.8 & 67.9 & 82.8  \\
Shared Embeddinging & 74.2 & 63.7 & 77.5  \\
MeMoE               & 75.8 & 65.2 & 79.4  \\
HMOE-Fuse           & 73.2 & 65.9 & 78.7  \\
Unified Modality Representation & 75.3 & 64.8 & 80.1  \\
\bottomrule
    \end{tabular}}
    \caption{Ablation studies on the unified embedding}
    \label{tab3}
\end{table}

\begin{table}[!http]
    
    \centering
\resizebox{0.95\linewidth}{!}{
\begin{tabular}{c|ccccc}
\toprule
\multirow{3}{*}{Method} & \multicolumn{1}{c}{\underline{SOT}} & 
\multicolumn{1}{c}{\underline{SOT}} & 
\multicolumn{1}{c}{\underline{VOS}} & 
\multicolumn{1}{c}{\underline{MOT}} & 
\multicolumn{1}{c}{\underline{MOTS}} \\
&GOT10K&LasHeR&LLE-VOS&UniRTL&BDD MOTS \\
&(AO)&(PR)&($\mathcal{J\&F}$)&HOTA&MOTSA\\ 
\midrule 
\rowcolor{purple!5} SATA & \textbf{81.3}&\textbf{77.8}&\textbf{71.4}&\textbf{59.7}&\textbf{38.1}  \\ 
\midrule   
Separate models       &78.5&74.3&68.7&56.4&36.3  \\
Task-specific heads   &80.8&75.4&67.5&54.2&35.5 \\ 
Multi-stage training  &79.5&75.2&68.4&57.3&37.6 \\
\midrule
\end{tabular}}
\caption{Ablation studies on the training strategy.}
\label{Tab12}
\end{table}

\textit{Training Strategy}
To investigate the effectiveness of the proposed TaMOT and our unified training strategy, we implement several comparative variants using existing methods, including:
\ding{172}: Training separate models for the SOT\&VOS task and MOT\&MOTS task, respectively.
\ding{173}: Employing task-specific heads.
\ding{174}: Adopting the multi-stage training strategy from \cite{unicorn}.
These methods are evaluated across 5 benchmarks (GOT10K~\cite{got10k}, LasHeR~\cite{lasher}, LLE-VOS~\cite{llevos}, UniRTL~\cite{unirtl}, and BDD MOTS~\cite{bdd100k}) covering 4 subtasks (SOT, VOS, MOT, MOTS).

Compared with our unified model, single-task models exhibit inferior performance across all tasks. Experimental results confirm that the proposed TaMOT can leverage expanded training data and further enhance the model's generalization capability through multi-task joint training.

\section{Limitation}
The current SATA is dedicated to constructing an effective framework for tracking and segmentation with any modality input, but it pay less attention on the efficiency of the proposed model. When tracking and segmenting multiple targets, the proposed method adopts a strategy of sharing unified embedding but tracking and segmenting these objects separately. Such a manner lacks interaction between objects and affects the efficiency of the model to some extent.

\end{document}